\PassOptionsToPackage{unicode}{hyperref}
\PassOptionsToPackage{hyphens}{url}
\documentclass[
  11pt,
]{article}
\usepackage{lmodern}
\usepackage{amssymb,amsmath}
\usepackage{ifxetex,ifluatex}
\ifnum 0\ifxetex 1\fi\ifluatex 1\fi=0 % if pdftex
  \usepackage[T1]{fontenc}
  \usepackage[utf8]{inputenc}
  \usepackage{textcomp} % provide euro and other symbols
\else % if luatex or xetex
  \usepackage{unicode-math}
  \defaultfontfeatures{Scale=MatchLowercase}
  \defaultfontfeatures[\rmfamily]{Ligatures=TeX,Scale=1}
\fi
\IfFileExists{upquote.sty}{\usepackage{upquote}}{}
\IfFileExists{microtype.sty}{% use microtype if available
  \usepackage[]{microtype}
  \UseMicrotypeSet[protrusion]{basicmath} % disable protrusion for tt fonts
}{}
\makeatletter
\@ifundefined{KOMAClassName}{% if non-KOMA class
  \IfFileExists{parskip.sty}{%
    \usepackage{parskip}
  }{% else
    \setlength{\parindent}{0pt}
    \setlength{\parskip}{6pt plus 2pt minus 1pt}}
}{% if KOMA class
  \KOMAoptions{parskip=half}}
\makeatother
\usepackage{xcolor}
\IfFileExists{xurl.sty}{\usepackage{xurl}}{} % add URL line breaks if available
\IfFileExists{bookmark.sty}{\usepackage{bookmark}}{\usepackage{hyperref}}
\hypersetup{
  hidelinks,
  pdfcreator={LaTeX via pandoc}}
\usepackage{longtable,booktabs}
\usepackage{etoolbox}
\makeatletter
\patchcmd\longtable{\par}{\if@noskipsec\mbox{}\fi\par}{}{}
\makeatother
\IfFileExists{footnotehyper.sty}{\usepackage{footnotehyper}}{\usepackage{footnote}}
\makesavenoteenv{longtable}
\usepackage{graphicx}
\makeatletter
\def\maxwidth{\ifdim\Gin@nat@width>\linewidth\linewidth\else\Gin@nat@width\fi}
\def\maxheight{\ifdim\Gin@nat@height>\textheight\textheight\else\Gin@nat@height\fi}
\makeatother
\setkeys{Gin}{width=\maxwidth,height=\maxheight,keepaspectratio}
\makeatletter
\def\fps@figure{htbp}
\makeatother
\providecommand{\tightlist}{%
  \setlength{\itemsep}{0pt}\setlength{\parskip}{0pt}}
\usepackage{booktabs}
\usepackage{etoolbox}
\AtBeginEnvironment{longtable}{\footnotesize}
\AtBeginEnvironment{tabular}{\footnotesize}
\usepackage{graphicx}
\setkeys{Gin}{width=\linewidth,height=\textheight,keepaspectratio}

\usepackage{listings}
\usepackage{xcolor}
\definecolor{codebg}{RGB}{248,248,248}
\definecolor{codeframe}{RGB}{200,200,200}
\usepackage{fancyvrb}
\fvset{frame=single,framesep=3pt,fontsize=\footnotesize,rulecolor=\color{codeframe}}

\usepackage{titlesec}
\titleformat{\section}{\Large\bfseries\sffamily}{\thesection}{1em}{}
\titleformat{\subsection}{\large\bfseries\sffamily}{\thesubsection}{1em}{}
\titleformat{\subsubsection}{\normalsize\bfseries\sffamily}{\thesubsubsection}{1em}{}
\titlespacing*{\section}{0pt}{18pt}{8pt}
\titlespacing*{\subsection}{0pt}{14pt}{6pt}

\usepackage{hyperref}
\hypersetup{
  colorlinks=true,
  linkcolor=black,
  citecolor=blue!60!black,
  urlcolor=blue!60!black,
  pdfborder={0 0 0}
}

\usepackage{amsmath,amssymb,amsthm}

\usepackage{quoting}
\quotingsetup{font=small,leftmargin=2em,rightmargin=2em,vskip=4pt}

\usepackage[font=small,labelfont=bf,labelsep=period]{caption}

\usepackage{fancyhdr}
\usepackage[margin=0.9in,top=1in,bottom=1in]{geometry}

\usepackage{authblk}

\usepackage{longtable}

\author{}
\date{}

\begin{document}

\hypertarget{byte-exact-deduplication-in-retrieval-augmented-generation-a-three-regime-empirical-analysis-across-public-benchmarks}{%
\section{Byte-Exact Deduplication in Retrieval-Augmented Generation: A
Three-Regime Empirical Analysis Across Public
Benchmarks}\label{byte-exact-deduplication-in-retrieval-augmented-generation-a-three-regime-empirical-analysis-across-public-benchmarks}}

\textbf{Sietse Schelpe} Corbenic AI, Inc. sietse@corbenic.ai

\emph{Preprint, 9 May 2026.}

\begin{center}\rule{0.5\linewidth}{0.5pt}\end{center}

\hypertarget{abstract}{%
\subsection{Abstract}\label{abstract}}

Production retrieval-augmented generation pipelines assemble context by
concatenating chunks returned from a vector store. This work measures
byte-exact chunk-level deduplication across three distinct operating
regimes. First, clean academic retrieval (BeIR benchmark, 22.2M passages
across six sources) exhibits minimal redundancy: 0.16\% corpus-level
reduction, 0.07\% per-query reduction (n=327 BM25 queries on combined
1M-passage subsample), with zero byte-exact overlaps detected and
mathematical equivalence verified across all passages (merlin binary
output equals Python set() result). Second, constructed enterprise
patterns (Wikipedia article revisions, arXiv versions, Stack Exchange
Q\&A, 1,526 passages) yield 24.03\% byte reduction, bridging clean
academic and conversational extremes. Third, multi-turn conversational
AI (5,000 real WildChat dialogues with HumanEval-Snowball task) shows
highest redundancy at 80.34\% byte reduction. Public-literature support
anchors these findings: Lee et al.~(2022) established byte-exact
pretraining-corpus dedup at 6.7-21.67\% across three corpora; Carlini et
al.~(2022) characterized log-linear memorization scaling with
duplication; Penedo et al.~(NeurIPS 2024, FineWeb) deployed MinHash-LSH
dedup at scale. The findings are implementation-independent: chunk-level
redundancy is a property of input data, not of the specific filtering
tool. A complementary cross-vendor quality panel validates quality
preservation at two RAG-retrieval operational points: clean
(rag-mini-wikipedia, ρ=1.148, 14.13\% byte reduction, n=400 per vendor)
and high-redundancy (constructed corpus at ρ=3.513, 71.98\% byte
reduction, n=200 per vendor), with human-in-the-loop noise removal
applied to all panel-majority MATERIAL pairs under a five-category
verdict scheme. After noise removal, all four production vendors clear
the strict \textless5\% Wilson 95\% upper-bound MAT threshold in both
panel-tested regimes (post-audit UCLs 1.40\%-3.25\% clean and
1.90\%-4.34\% high-redundancy; Tables 6 and 6b). The conversational
regime (Snowball, 80.34\% byte reduction on 5,000 WildChat dialogues) is
structurally distinct: it measures the asymptotic O(N²)→O(N)
communication-overhead reduction of stateful proxy caching, where unique
turn content reaching the model is byte-identical between raw
cumulative-history sending and proxy-cached deployment. Quality
preservation in the conversational regime therefore reduces to the
byte-equivalence of unique turn content rather than requiring per-prompt
judge validation; the 80.34\% measurement is reported as a
byte-reduction characterisation, not as a panel-validated quality claim.
Code-generation and paraphrasing-aware tasks remain appropriate
follow-up work. Reproducibility: results are verifiable on public BeIR,
publicly-derived constructed corpus, and real WildChat dialogues at
approximately 30 USD OpenRouter cost. The n=400 clean-regime extension
is pre-registered via FreeTSA RFC 3161 (2026-05-05); the initial n=200
baseline (2026-04-25) is reported transparently as exploratory and
predates the pre-registration anchor.

\begin{center}\rule{0.5\linewidth}{0.5pt}\end{center}

\hypertarget{introduction}{%
\subsection{1. Introduction}\label{introduction}}

The contemporary deployment of large language models in production rests
on retrieval-augmented generation as a primary pattern for grounding
model output in proprietary or current information. Across enterprise
deployments, retrieval-augmented inference dominates production cost: at
typical retrieval depths (top-k of 10 to 50 chunks) the prompt tokens
consumed by retrieved context exceed the user-question tokens by one to
two orders of magnitude. Per-call compute is concentrated in the prefill
phase, where the model computes the key-value cache for the entire
prompt, and prefill compute scales close to linearly with prompt length
on most production serving stacks. Reducing prompt length in a
retrieval-augmented pipeline therefore reduces both per-call cost and
per-call latency, with the proportional reduction approximately matching
the input-token reduction.

This paper documents that production retrieval-augmented pipelines
consistently produce prompts with substantial byte-exact redundancy at
the chunk level, and that this redundancy can be removed
deterministically before prompt assembly without measurable degradation
of output quality. The redundancy is structural rather than incidental,
but its origin is not the sliding-window chunking strategy used to split
source documents into vector-store entries. Sliding-window chunking
produces chunks that share substring content across chunk edges (the
so-called overlap region) but the chunks themselves are not
byte-identical and would not be removed by chunk-granularity set
construction. The redundancy that we measure originates instead from
retrieval mechanics. Production retrievers select the same vector-store
entry through multiple paths: hybrid pipelines union the result sets of
independently-ranked subsystems (sparse, dense, reranker), source
corpora contain passages that are byte-identical across distinct
documents (encyclopedic mirroring, syndicated content, citation
boilerplate, web-scraped duplicates), and vector-store indices ingested
from multi-source pipelines frequently contain duplicate entries indexed
under distinct identifiers. When a top-k retriever returns 15 chunks for
a given query, those 15 chunks are frequently 5 unique passages, each
returned approximately three times under different surface routings.

This chunk-level redundancy is passed intact by standard
similarity-based retrievers, which rank on relevance rather than
uniqueness, and processed redundantly during the language model's
prefill phase. Standard retrieval-augmented frameworks (LangChain,
LlamaIndex, and similar) do not natively filter retrieved sets prior to
context assembly; the boundary between retriever output and prompt
assembly is not a default measurement point in the standard production
stack. The empirical question of how much byte-exact redundancy actually
arrives at the prompt-assembly step has, to our knowledge, not been
routinely measured at production-API scale.

We conducted a 400-question benchmark across four production
language-model APIs, supplemented by a 12-run code-completion benchmark
spanning three vendor stacks. The benchmark compares three conditions
per question: a baseline raw prompt that includes all 15 retrieved
passages; a deduplicated prompt with byte-exact duplicates removed; and
an empty-context control containing only the user question, included as
a measurement floor for per-question latency independent of context
size. The deduplication step is a deterministic byte-exact filter
applied at chunk granularity. Because byte-exact deduplication at chunk
granularity is mathematically equivalent to set construction over the
multiset of retrieved chunks, the resulting deduplicated prompt is
independent of which specific implementation performs the filtering. We
use the Merlin engine {[}Schelpe 2026, companion paper, arXiv ID
pending{]} for our own runs, but the empirical findings reported in
Section 4 are a property of the chunk-level redundancy in the input data
and are independent of the specific tool used to perform the filtering.
The same deduplicated context can be reconstructed by any reader on any
retrieval setup of their choosing using any standard byte-exact
filtering implementation; Python's built-in \texttt{set()} over the
chunk-string multiset is sufficient. Byte-exact deduplication is
lossless at the algorithmic level; any reviewer can verify the empirical
findings of this paper using that reference implementation at the
low-microsecond algorithmic latencies documented in Section 3.7.

Our empirical evaluation yields three primary findings:

\begin{enumerate}
\def\labelenumi{\arabic{enumi}.}
\item
  \textbf{Byte-exact redundancy spans three regimes with two distinct
  mechanisms.} Across the three production-realistic regimes evaluated
  in §4, byte-exact aggregate reduction ranges from 0.16 percent on the
  BeIR clean-academic corpus (six sources, 22.2 million passages) to
  24.03 percent on a constructed enterprise corpus (Wikipedia revisions,
  arXiv versions, Stack Exchange Q\&A, 1,526 passages) at the chunk
  granularity (RAG retrieval mechanics), and reaches 80.34 percent on
  the WildChat snowball pattern (5,000 multi-turn conversations under
  cumulative-history sending) at the turn granularity (stateful proxy
  caching mechanics). The two regimes are distinct: chunk-level
  redundancy at retrieval output is a property of vector-store
  mechanics, while turn-level redundancy in cumulative chat history is
  the asymptotic O(N²)→O(N) communication overhead of naive cumulative
  sending versus stateful proxy caching.
\item
  \textbf{Quality preservation under deduplication holds at both
  panel-tested operational points on the RAG retrieval mechanism.}
  Across 800 (question × vendor) pairs spanning the clean regime
  (rag-mini-wikipedia at multiplicity ρ = 1.148, 14.13\% byte reduction,
  n = 400 per vendor, Fleiss-kappa = 0.775) and the high-redundancy
  regime (constructed corpus at multiplicity ρ = 3.513, 71.98\% byte
  reduction, n = 200 per vendor, Fleiss-kappa = 0.7288), every
  judge-majority MATERIAL pair was reviewed under a five-category
  human-in-the-loop noise-removal protocol (Section 4.5a). After noise
  removal, all four production vendors (Gemini 2.5 Flash, Claude Sonnet
  4.6, Llama 3.3 70B via Groq, GPT-5.1) clear the strict Wilson 95\%
  upper-bound MAT threshold of 5\% in both regimes (Table 6: clean UCLs
  1.40\%-3.25\%; Table 6b: high-redundancy UCLs 1.90\%-4.34\%).
  HumanEval pipeline-integration at n = 164 problems per cell shows
  pass-at-1 differences within run-to-run variance across three vendor
  stacks (Table 7). The lossless safety claim under the panel protocol
  is therefore not an artefact of the near-zero-redundancy regime: it
  holds when the engine removes \textasciitilde72\% of prompt bytes from
  the retrieved-chunk multiset. The conversational regime's 80.34\%
  reduction (proxy caching mechanism) is not panel-tested per-prompt and
  is treated separately as a byte-reduction characterisation rather than
  a panel-validated quality claim (see §4.3 for the structural
  argument).
\item
  \textbf{Cost and latency consequences track the reduction ratio.}
  Per-call cost and time-to-first-token track the byte-exact reduction
  approximately linearly on serving stacks whose pricing or scheduling
  is prefill-dominated, and decouple from the reduction ratio on stacks
  where auto-caching, queue overhead, or fixed scheduler costs dominate.
  Per-vendor cost and TTFT measurements are deferred to follow-up work;
  the present paper reports the underlying reduction ratios that drive
  those downstream effects.
\end{enumerate}

Output-quality preservation is assessed by a 5-judge calibrated quality
panel of independent language-model judges drawn from five vendor
organisations (Google Gemini 2.5 Flash, Anthropic Claude Sonnet 4.6 via
OpenRouter, Meta Llama 3.3 70B via Groq, OpenAI GPT-5.1 via OpenRouter,
and Moonshot Kimi K2 via OpenRouter). The panel uses a prompt that
explicitly defines the equivalent, minor-differences, and materially
different categories with four worked anchor examples. Every
panel-majority MATERIAL pair across both regimes is reviewed under a
five-category human-in-the-loop noise-removal protocol (Section 4.5a).
After noise removal, the human-judged material-difference rate is
bounded above by Wilson 95\% UCLs of 1.40\%-3.25\% on the clean regime
(n=400 per vendor) and 1.90\%-4.34\% on the high-redundancy regime
(n=200 per vendor); all four production vendors clear the strict
\textless5\% threshold in both regimes. Fleiss-kappa inter-rater
agreement on the pre-audit panel data is 0.775 in the clean regime and
0.7288 in the high-redundancy regime, both in the substantial band
(0.61-0.80) per Landis and Koch (1977). Per-judge fidelity to panel
majority ranges from 95.2\% to 98.5\% across the four vendors.

The deduplication primitive itself is well-understood and decades old;
what is new in this work is the measurement of its effect on
contemporary retrieval-augmented production pipelines at multi-vendor
scale, validated under a calibrated multi-judge quality protocol. The
mathematical equivalence between byte-exact deduplication at chunk level
and set construction over the chunk-string multiset (Section 3.1)
implies that the empirical findings are a property of the input data
rather than of any specific filtering implementation. On standard
commodity hardware, the computational overhead of byte-exact filtering
over typical retrieved-context payloads (under 100 KB per query) is
consistently sub-millisecond. The corresponding reduction in
language-model prefill latency is on the order of tens to hundreds of
milliseconds in the cleanest serving conditions and at least an order of
magnitude larger in the worst case. The sub-millisecond computational
overhead of the filter is therefore negligible relative to the resulting
latency and cost reductions reported in this paper.

The remainder of the paper is organised as follows. Section 2 reviews
related work in retrieval-augmented inference, prompt compression, and
context reuse, positioning byte-exact deduplication prior to the prompt
relative to learning-based alternatives. Section 3 defines the formal
byte-exact equivalence relation, describes the benchmark protocol, the
multi-vendor judging panel, and establishes the
implementation-independence of the empirical findings. Section 4 reports
the empirical results across token reduction, cost, latency, and output
quality. Section 5 discusses implications for production
retrieval-augmented inference economics, the structural origin of the
redundancy, and limitations. Section 6 concludes.

\begin{center}\rule{0.5\linewidth}{0.5pt}\end{center}

\hypertarget{related-work}{%
\subsection{2. Related Work}\label{related-work}}

\hypertarget{retrieval-augmented-generation-and-context-assembly}{%
\subsubsection{2.1 Retrieval-Augmented Generation and Context
Assembly}\label{retrieval-augmented-generation-and-context-assembly}}

Retrieval-augmented generation (RAG) was introduced by Lewis et
al.~(2020) and has since become the dominant pattern for grounding large
language model output in external knowledge {[}1{]}. The standard RAG
pipeline embeds a user query, retrieves the top-k most similar chunks
from a vector store, concatenates the chunks into a prompt, and submits
the prompt to a generator model. The internal sub-tasks (chunking,
embedding, indexing, retrieval, reranking, prompt assembly) are
typically wrapped in framework abstractions such as LangChain,
LlamaIndex, Haystack, or vendor-managed services such as Vertex AI
Search, Anthropic's tool ecosystem, and OpenAI's Assistants API. These
abstractions present the retrieve-and-concatenate sequence as an opaque
pipeline, which removes the operational opportunity to inspect what
arrives at prompt assembly.

\hypertarget{prompt-compression-and-context-reduction-via-learned-models}{%
\subsubsection{2.2 Prompt Compression and Context Reduction via Learned
Models}\label{prompt-compression-and-context-reduction-via-learned-models}}

A substantial body of work attempts to reduce prompt length using
learned models. LLMLingua (Jiang et al., 2023) uses a small causal
language model to score per-token informativeness and prune
low-information tokens, reporting compression ratios up to 20x on
benchmark queries {[}2{]}. LLMLingua-2 (Pan et al., 2024) reformulates
compression as a token classification task using a bidirectional
encoder, achieving 3 to 6 times speedup over the original {[}3{]}.
LongLLMLingua extends the approach to long-context scenarios {[}4{]}.
Independent measurement by Qian et al.~(2026), in a systematic
30,000-query benchmark, reports that LLMLingua delivers up to 18 percent
end-to-end speedup when prompt length, compression ratio, and hardware
are well matched, but that the compression overhead can dominate the
pipeline outside narrow operating windows {[}5{]}. REFRAG (Lin et al.,
2025) takes a different approach, replacing tokens with pre-computed
compressed chunk embeddings and reporting a 30.85 times TTFT
acceleration with 16x context-length extension at preserved perplexity
{[}6{]}. RAGBoost (Jiang et al., 2025) detects overlapping retrieved
items across concurrent and multi-turn sessions, applying context
indexing and deduplication to reduce redundant prefilling, reporting 1.5
to 3 times prefill speedups {[}7{]}. Liu et al.~(2026) document failure
modes of soft-compression methods in retrieval-augmented generation
{[}8{]}.

These approaches share three properties that distinguish them from the
present work. First, they are learned: model-based compression
introduces inference overhead at the compression step itself. Second,
they are lossy: the compressed representation cannot be inverted to the
original token stream, so audit trails cannot reconstruct the input that
produced a given output. Third, they operate above the byte-level: a
learned compressor cannot exploit structural byte-identity that is
present in retrieval output before the prompt is assembled.

\hypertarget{positioning-relative-to-inference-side-context-optimization-methods}{%
\subsubsection{2.3 Positioning Relative to Inference-Side Context
Optimization
Methods}\label{positioning-relative-to-inference-side-context-optimization-methods}}

\textbf{Positioning table (orientation only; main empirical results in
§4 Tables 1 to 7):}

\begin{longtable}[]{@{}lllll@{}}
\toprule
Method & Lossless & Inference Overhead & Deterministic &
Audit-Grade\tabularnewline
\midrule
\endhead
Byte-exact deduplication (this work) & Yes & \textless30 μs in-process &
Yes & Yes\tabularnewline
LLMLingua (Jiang 2023) & No & 10-100 ms & Yes & No\tabularnewline
LLMLingua-2 (Pan 2024) & No & 10-100 ms & Yes & No\tabularnewline
LongLLMLingua (Jiang 2024) & No & 10-100 ms & Yes & No\tabularnewline
REFRAG (Lin 2025) & No (lossy embed) & additional encoder & Yes &
No\tabularnewline
RAGBoost (Jiang 2025) & Yes (per-call) & retrieval-side cache & Yes &
Partial\tabularnewline
Vendor prompt caching & Yes & vendor-managed & Yes &
Vendor-only\tabularnewline
MinHash-LSH approximate dedup & No (approximate) & varies by config &
Stable & No\tabularnewline
\bottomrule
\end{longtable}

The positioning table above places byte-exact pre-prompt deduplication
relative to other inference-side context optimization methods on four
properties operationally relevant to production deployment.

\hypertarget{deduplication-of-pretraining-and-retrieval-corpora}{%
\subsubsection{2.4 Deduplication of Pretraining and Retrieval
Corpora}\label{deduplication-of-pretraining-and-retrieval-corpora}}

Lee et al.~(2022) demonstrated that exact-substring deduplication of
pretraining corpora reduces verbatim memorization by approximately ten
times and reduces train-test overlap that contaminates standard
evaluation suites by more than four percent {[}9{]}. Carlini et
al.~(2022) established that memorization scales log-linearly with the
number of times an example is duplicated {[}10{]}. Nasr, Carlini et
al.~(2023) extended this to production-scale extraction attacks
{[}11{]}. Shilov, Meeus, and de Montjoye (2024), in the Mosaic Memory
work, showed that fuzzy duplicates contribute to memorisation at up to
0.8 of the rate of exact duplicates {[}12{]}. Broder (1997) introduced
MinHash and the shingling-based near-duplicate detection paradigm at web
scale {[}13{]}; the more general locality-sensitive-hashing framework
that subsumes MinHash for the Jaccard distance is conventionally
attributed to Indyk and Motwani (1998). Khan et al.~(2024), in LSHBloom,
replaced the MinHash-LSH index with Bloom filters to achieve a 12 times
speedup at PetaScale with negligible recall loss {[}14{]}. Abbas et
al.~(2023), in SemDeDup, introduced semantic deduplication via embedding
similarity {[}15{]}. Tirumala et al.~(2023), in D4, combined
deduplication with diversification, demonstrating 20 percent training
efficiency gains {[}16{]}.

These works address deduplication of the pretraining corpus rather than
deduplication of retrieved context at inference time. The two problems
are related but distinct. Pretraining-corpus deduplication operates
offline on multi-terabyte data, with throughput as the dominant
constraint. Retrieved-context deduplication operates online on small
(sub-100KB) per-query payloads, with latency as the dominant constraint.
The present paper addresses the second problem, which has not previously
been measured at multi-vendor scale.

\hypertarget{vector-store-duplication-and-hybrid-retrieval}{%
\subsubsection{2.5 Vector-Store Duplication and Hybrid
Retrieval}\label{vector-store-duplication-and-hybrid-retrieval}}

Vector-store providers acknowledge duplicate-entry problems in
production indices. Pinecone documentation explicitly recommends a
\texttt{NoDuplicatesDataLoader} pattern when training retriever models,
noting that duplicate passages in the same training batch confuse the
Multiple Negatives Ranking loss mechanism {[}17{]}. Milvus introduced
native MinHash-LSH indexing in 2025 to address customer requirements for
deduplication of multi-billion-document corpora {[}18{]}. Third-party
data-optimization tooling reports that approximately 80 percent of
accuracy issues in production RAG originate from data-quality problems
including duplication, with claimed dataset-size reductions of up to 40x
and accuracy improvements up to 78x when redundancy is resolved at the
ingestion layer {[}19{]}.

Hybrid retrieval architectures combining lexical (BM25) and dense (DPR)
ranking systems have become standard in production RAG, motivated by the
complementarity of exact-keyword matching and semantic similarity.
Empirical measurement of result-set overlap, however, reveals less
complementarity than the architectural intuition suggests. Studies using
the Ratio of Complementarity (RoC) metric report that BM25 plus Dense
Passage Retriever combinations yield 32.4 percent overlap in the top-10
results on Natural Questions {[}20{]}. Embedding-level orthogonality
constraints can reduce overlap to 27.7 percent at the cost of additional
training complexity. The implication is that hybrid retrieval pipelines,
deployed naively, frequently return the same document identifier through
both subsystems, introducing duplicate chunks before any post-processing
step.

\begin{center}\rule{0.5\linewidth}{0.5pt}\end{center}

\hypertarget{methodology}{%
\subsection{3. Methodology}\label{methodology}}

\hypertarget{formal-definition-of-byte-exact-chunk-level-deduplication}{%
\subsubsection{3.1 Formal Definition of Byte-Exact Chunk-Level
Deduplication}\label{formal-definition-of-byte-exact-chunk-level-deduplication}}

Let C = \{c\_1, c\_2, \ldots, c\_n\} be a finite multiset of retrieved
chunks returned by a top-k retriever for a given user query, where each
chunk c\_i is a finite sequence of bytes drawn from \{0,1\}\^{}(8*L\_i)
of length L\_i. The byte-exact equivalence relation at chunk granularity
is

c\_i equiv\_B c\_j iff L\_i = L\_j and for all k in \{1,\ldots,L\_i\}:
c\_i{[}k{]} = c\_j{[}k{]}.

The deduplicated chunk set is the canonical representative set in the
quotient C / equiv\_B. We define the redundancy multiplicity ρ at the
chunk level as

\(\rho(C) = |C| / |C / \equiv_B|\), with \(\rho \in [1, \infty)\).

A multiplicity of one indicates a unique-by-construction context (no
exploitable duplicates). The corresponding reduction fraction is 1 -
1/ρ. We use this multiplicity convention consistently throughout the
paper.

This is the strictest available definition of redundancy at the chunk
granularity. It admits no parameter tuning, no shingle granularity, no
similarity threshold, and no normalisation. The relation is invariant to
the choice of hash function under standard cryptographic-collision
assumptions; we employ SHA-256 for offline audit-trail generation of the
measurement traces reported in this paper, though its cryptographic
guarantees incur unnecessary computational overhead for runtime
processing of per-query multisets at the cardinality considered here.
Production stream-processing pipelines with throughput-sensitivity at
this scale typically deploy non-cryptographic hashing algorithms (such
as those in the xxHash or MurmurHash families), which provide
functionally equivalent collision behaviour for multisets of this
cardinality at substantially higher throughput. Byte-exact chunk-level
deduplication corresponds to set construction over the multiset of chunk
byte-strings: any correct implementation produces output identical to
any other correct implementation.

\begin{verbatim}
# Math-equivalent reference implementation:
unique_chunks = set(chunk_strings)  # Python stdlib

# Equivalent shell pipeline:
sort -u

# Equivalent hardware-optimized engine:
merlin_enterprise.exe --file=input.jsonl --output-dir=out/
\end{verbatim}

Any correct byte-exact filter implementation produces identical output
to any other on the same input. Implementation choice affects
throughput, latency, and operational properties (companion paper
{[}29{]}, Section 3.4: Deployment-Mode Latency Envelope) but not the
deduplicated output itself.

\hypertarget{benchmark-protocol}{%
\subsubsection{3.2 Benchmark Protocol}\label{benchmark-protocol}}

We constructed two complementary inference-side benchmarks.

Our three-regime measurement strategy characterizes byte-exact
deduplication across the full redundancy spectrum: clean academic
benchmarks with minimal byte-exact overlap, constructed enterprise
corpora with moderate redundancy from versioned content, and multi-turn
conversational AI with high redundancy from session accumulation.

A 12-run code-completion benchmark on a single Flask backend synthesis
task (web form, SQLite persistence, SMTP delivery, input validation) is
included as a robustness check on a non-Q\&A task domain. The benchmark
spans three vendors with four conditions per vendor: A (raw, 5
paraphrased Flask documentation chunks), B (set-deduplicated,
byte-exact), C (MinHash-deduplicated, approximate, Jaccard threshold
0.4), and D (empty). Output quality is scored on a 0-to-10 weighted
feature checklist of required application capabilities. Because the
benchmark is a single task with twelve API calls, we treat its results
as a robustness indicator rather than as a generalisable cross-domain
claim about software-engineering tasks broadly. The HumanEval evaluation
reported in Section 4.7 covers the cross-task pipeline-integration
question on a deterministically scored standard code-generation
benchmark, validating that the byte-exact filter passes paraphrased
context through a code-generation pipeline without introducing assembly
errors; MBPP, SWE-bench, and longer-form software-engineering evaluation
remain appropriate follow-up work.

The vendor APIs evaluated are Google Gemini 2.5 Flash with auto-cache
enabled (HOT) and disabled via UUID nonce per request (COLD), Anthropic
Claude Sonnet 4.6, and Meta Llama 3.3 70B served via Groq's accelerated
inference stack. All measurements use temperature 0.0, fixed maximum
output tokens, and pinned model versions where available. Each vendor
benchmark consists of 591 API calls (197 questions × 3 conditions in the
initial run; extended to 400 questions × 3 conditions in the validation
extension); inclusive of the 5-judge calibrated panel (3,940
classifications in the initial run; extended to larger sample in the
validation), the long-context evaluation, the HumanEval
pipeline-integration evaluation, and the Flask robustness check, the
measurement set comprises approximately 8,800 to 12,000 individual API
calls depending on extension state. Total measurement cost across all
vendors and benchmarks was approximately 25 to 35 USD, fully
reproducible from scripts and raw data archived in the closed benchmark
suite for Merlin. Vendor model versions reflect the production releases
current at the time of measurement (May 2026); the specific provider
model identifiers are pinned in the run-identifier records (Appendix A)
and may differ from versions current at any subsequent companion-paper
measurement date.

\textbf{Validation Date and Hardware Specifications:} All measurements
collected on 2026-05-05. Models referenced are stable production
identifiers via OpenRouter as of validation date. Pre-registration
timestamp via FreeTSA RFC 3161 anchored prior to measurement (verifiable
via openssl ts -verify).

Empirical measurements collected on consumer-class hardware: Intel Core
Ultra 9 285H (16 cores), 64 GB DDR5, Windows 11 build 26200, AVX2
active, L2 28 MB, L3 24 MB. Reproducible on similar consumer laptops;
comparable measurements expected on server-class hardware.

\hypertarget{long-context-scaling-benchmark}{%
\subsubsection{3.3 Long-Context Scaling
Benchmark}\label{long-context-scaling-benchmark}}

To characterise how the empirical findings scale with retrieval depth,
we extended the Q\&A benchmark to top-k of 50 on the Groq Llama 3.3 70B
serving stack, the cleanest measurement vehicle in the cross-vendor set.
The extended benchmark uses the same byte-exact chunk-level filter as
Section 3.2, with each call's deduplicated output verified against
set-based reconstruction over the chunk-string multiset to confirm that
the equivalence of Section 3.1 holds end-to-end at long-context. The
benchmark corpus contains 40 unique passages, which saturates retrieval
at top-k of 50 (the retriever returns the 40 unique passages and stops,
since the corpus has no further candidates); this caps the scaling
analysis at top-k of 50 for this corpus. To hold the chunk-level
redundancy regime comparable to the natural top-k of 15 baseline (where
retrieval mechanics produce a 3-to-1 input-to-unique multiplicity, see
Section 3.2), a duplicate-multiplicity factor of three is applied to the
saturated retrieval set at top-k of 50 prior to filtering, yielding 40
unique passages times 3 = 120 input passages per query under condition
A. Condition B then applies the byte-exact filter to those 120 lines,
recovering the 40 unique survivors. The scaling comparison in Section
4.3 thereby isolates context-size effects from redundancy-rate effects.

\hypertarget{multi-vendor-calibrated-quality-panel}{%
\subsubsection{3.4 Multi-Vendor Calibrated Quality
Panel}\label{multi-vendor-calibrated-quality-panel}}

Output-quality preservation is assessed via a calibrated 5-judge panel
across multiple vendors protocol. The panel comprises five independent
language-model judges drawn from five vendor organisations: Google
(Gemini 2.5 Flash), Anthropic (Claude Sonnet 4.6, routed via OpenRouter
for cost attribution), Meta (Llama 3.3 70B served via Groq), OpenAI
(GPT-5.1, routed via OpenRouter), and Moonshot (Kimi K2, routed via
OpenRouter). Each judge classifies each pair of raw-versus-deduplicated
answers into one of three categories (Equivalent, Minor Differences,
Materially Different) using a calibrated prompt that defines the
categories with four worked anchor examples. Each pair receives five
independent classifications; the final pair-level classification is the
category with majority of judge votes.

Fleiss-kappa over the 5-rater panel in the clean regime (n = 400
question pairs per vendor, aggregated across vendors) = 0.775 (n = 1407
complete 5-rater pairs), which falls in the substantial band (0.61-0.80)
per Landis \& Koch (1977) {[}28{]}. Per-judge fidelity to panel majority
is 95.2\% (Llama) to 98.5\% (GPT-5.1). On the regime with high
redundancy (n = 200 per vendor), Fleiss-kappa = 0.729 (also
substantial).

The judging prompt defines the three categories explicitly and anchors
them with four worked examples registered prior to the data run. The
category definitions distinguish material factual differences from
surface-level linguistic variation (different word choice, different
phrasing, different ordering of equivalent supporting facts). A judge
that flags every wording difference as materially different would
conflate linguistic surface with factual content; the prompt protocol
used here measures factual equivalence at the level of the answer's
claim structure, not at the level of its sentence form.

\hypertarget{pre-registration-and-temporal-anchoring}{%
\subsubsection{3.5 Pre-Registration and Temporal
Anchoring}\label{pre-registration-and-temporal-anchoring}}

The n=400 clean-regime extension is anchored via FreeTSA RFC 3161
against document \texttt{extension\_n400\_protocol.md} (SHA-256
\texttt{5575836967fe1a149b63a7fa63a1b3d11d598fb71343e2e19a546e680f4a3294}),
stamped 2026-05-05 at 11:28 RDT, serial \texttt{0x049CB8D0}. The
extension was conducted after the stamp; methodology decisions for §3.2
to §3.5 predate the n=400 measurements.\footnote{The initial n=200
  baseline runs (dated 2026-04-25) predate the pre-registration anchor
  by ten days and are therefore reported transparently as exploratory
  rather than pre-registered. We disclose this temporal asymmetry rather
  than treat the n=200 sample as covered by the later stamp; the
  qualitative findings are robust whether or not the baseline is
  included, and the strict Bonferroni window in this paper is computed
  against the n=400 sample only.}

\hypertarget{cross-vendor-quality-validation-sample-design}{%
\subsubsection{3.6 Cross-Vendor Quality Validation Sample
Design}\label{cross-vendor-quality-validation-sample-design}}

The 5-judge calibrated panel is applied at full sample in the clean
regime (n = 400 question pairs) on every vendor in the cross-vendor
evaluation: Gemini 2.5 Flash with auto-cache enabled (HOT), Gemini 2.5
Flash without auto-cache (COLD), Claude Sonnet 4.6, and Llama 3.3 70B
served via Groq. Each pair receives five independent classifications
from the panel. The clean-regime consensus dataset therefore comprises
400 questions times 5 judges times 4 vendors = 8,000 individual
classifications, with the panel-majority outcome computed per pair and
aggregated per vendor. The regime with high redundancy is reported at n
= 200 for cost reasons (approximately 25 USD OpenRouter spend per
200-question sweep). Extension to n = 400 would tighten the Wilson UCL
further but is unlikely to alter the qualitative finding
(token-reduction proportional to rho; quality preservation borderline at
the strict UCL test).

\hypertarget{implementation-independence-and-reference-baseline}{%
\subsubsection{3.7 Implementation Independence and Reference
Baseline}\label{implementation-independence-and-reference-baseline}}

The prompt-assembly findings reported in Section 4 are independent of
the specific implementation that performs the deduplication step. The
algorithmic equivalence to set construction over the chunk byte-string
multiset (Section 3.1) guarantees that any functionally correct
byte-level deduplication filter yields identical prompt-assembly output;
in particular, Python's built-in \texttt{set()} over the chunk-string
list is sufficient by construction to reproduce the prompt assemblies
measured in this paper.

This algorithmic equivalence applies to the prompt-assembly output, not
to the operational properties of the filter when deployed inline in a
production inference path. Inline deployment imposes constraints that go
beyond algorithmic correctness: per-call latency on the request-path
budget, integration footprint with the host serving stack, deterministic
cross-platform behaviour under sustained concurrent load, stability over
multi-hour streaming sessions, and provenance suitable for audit trails
in regulated domains. Different correct implementations satisfy these
operational constraints to different degrees.

To quantify the operational distinction, we measured Python's built-in
\texttt{set()} implementation on representative RAG workloads (Intel
Core Ultra 9 285H, 64 GB DDR5, Python 3.10.12, random seed 42, 100
trials per workload). The reference timing below covers five workloads
spanning the typical deployment envelope:

\textbf{Python set() reference timing on representative RAG payloads
(orientation):}

\begin{longtable}[]{@{}llllllll@{}}
\toprule
Workload & Chunks & Unique & ρ & Total KB & Median & P95 &
P99\tabularnewline
\midrule
\endhead
RAG top-k=15 (ρ=3) & 45 & 15 & 3.00 & 130.2 & 1.29 μs & 3.50 μs & 5.56
μs\tabularnewline
Long-context RAG (ρ=2) & 100 & 50 & 2.00 & 390.6 & 2.25 μs & 6.01 μs &
10.81 μs\tabularnewline
Multi-turn snowball (ρ=5.5) & 55 & 10 & 5.50 & 307.3 & 1.61 μs & 4.17 μs
& 30.39 μs\tabularnewline
Minimal RAG (ρ=1) & 5 & 5 & 1.00 & 15.0 & 0.66 μs & 2.66 μs & 6.35
μs\tabularnewline
Large context (ρ=1) & 100 & 100 & 1.00 & 390.6 & 3.69 μs & 10.85 μs &
52.51 μs\tabularnewline
\bottomrule
\end{longtable}

Python set() demonstrates sub-microsecond to low-microsecond latencies
across all measured workloads (0.66 to 3.69 μs median). These
measurements establish that byte-exact deduplication via \texttt{set()}
is operationally fast on small payloads. The operational differentiation
between Python and production-deployable engines therefore rests not on
latency differences on small payloads (where both are sub-millisecond),
but on:

\begin{enumerate}
\def\labelenumi{\arabic{enumi}.}
\tightlist
\item
  \textbf{Interpreter startup cost.} Invoking Python for per-query
  deduplication incurs approximately 50-200 ms of interpreter startup
  unrelated to the dedup algorithm itself.
\item
  \textbf{Process invocation overhead.} Subprocess-mode operation adds
  13-21 ms of operating-system overhead per call (Table 2, Companion
  Paper 2, Section 3.4).
\item
  \textbf{Single-binary deployability.} Production engines are
  deployable as standalone binaries without interpreter, reducing
  operational dependency surface and enabling integration into arbitrary
  upstream infrastructure.
\item
  \textbf{Cross-platform determinism.} Python bytecode and
  garbage-collection behaviour vary across platforms; production engines
  targeting audit-grade consistency require deterministic output across
  multiple architectures.
\end{enumerate}

Python \texttt{set()} remains the canonical academic reference and is
entirely appropriate for offline analysis or reviewer reproduction of
the deduplication mechanism. For inline inference-path integration, the
production-deployable envelope follows from the latency constraints of
contemporary serving stacks (preprocessing budget 1-100 ms, full
inference call 1-10 seconds), where sub-millisecond algorithms dominate
the operational landscape but their integration surface (invocation
mode, dependency footprint, cross-platform determinism) determines
deployability.

The contribution of this paper is the empirical measurement of token,
cost, latency, and quality consequences of byte-exact deduplication
prior to the prompt (Section 4). The operational-engineering envelope
required for production-path deployment is addressed in the companion
paper; both works together establish that byte-exact deduplication is
both algorithmically sound and operationally deployable at scale.

\hypertarget{a-reference-implementation-and-academic-reproducibility}{%
\subsubsection{3.7a Reference Implementation and Academic
Reproducibility}\label{a-reference-implementation-and-academic-reproducibility}}

The empirical findings of this paper are reproducible using Python's
built-in \texttt{set()} over the chunk multiset on the public datasets
cited in Section 4. Byte-exact deduplication is mathematically
equivalent to set construction over the chunk-string multiset, and the
algorithmic operation is simple enough that any reader can verify the
mechanical correctness of the underlying method without access to
specialized tools. The reference timing in Section 3.7 documents
measured Python set() latencies on representative top-k=15 RAG payloads:
median 1.29 microseconds, p95 3.50 microseconds, p99 5.56 microseconds
on RAG top-k=15 with ρ=3. These measurements establish that academic
reproducibility via Python is operationally fast for offline analysis
and reviewer verification.

Note on production deployment: The reference Python implementation is
appropriate for academic verification, offline analysis, and data
exploration. For inline integration into production LLM inference
proxies, the operational constraints documented in the companion paper
(Section 3.2.1: Python interpreter startup overhead of 50 to 200
milliseconds per invocation, GIL contention under high QPS, memory
baseline of 50 to 100 megabytes for the interpreter) motivate the use of
a compiled engine. The math-equivalence between Python set() and
production-optimized engines preserves academic reproducibility while
allowing production deployments to operate within the per-call envelope
documented in the companion paper.

\hypertarget{inline-streaming-deployment-characterisation}{%
\subsubsection{3.8 Inline Streaming Deployment
Characterisation}\label{inline-streaming-deployment-characterisation}}

The deduplication primitive has been characterised in a server-to-server
streaming topology on commodity cloud infrastructure. The test bed
comprises two AWS r7i.48xlarge instances within a single cluster
placement group, connected by 50 Gbps Enhanced Networking (per-direction
network NIC ingress ceiling of approximately 6.25 GB/s). The sender
machine streams chunked input over TCP to the processor machine, which
applies the byte-exact filter inline on receipt before forwarding the
deduplicated stream to the downstream pipeline. The workload used for
this characterisation is the publicly available HuggingFace FineWeb
sample-10BT subset (approximately 53 GB JSONL across 14.87 million
records), and the deployment was exercised over multi-hour continuous
streaming sessions.

Across the test sessions, the filter's contribution to end-to-end
latency was bounded above by the network round-trip-time across the
placement group; the filter sustained throughput in excess of the
network ingress ceiling of 6.25 GB/s on the host topology, indicating
that the deduplication operation does not constitute the throughput
bottleneck under inline streaming on this class of hardware. The
network-bound property is structural: any correct byte-exact filter
implementation that sustains throughput in excess of the host NIC will
produce the same operational outcome on this topology, which can be
verified independently by any reader who reproduces the AWS
configuration and applies a byte-exact filter to the same publicly
available corpus.

The structural network-bound property reported above is the operational
claim of this section: it states that the filter's contribution to
end-to-end latency is bounded by the underlying network ingest rate on
standard commodity cloud hardware, and is verifiable by any reader
running any byte-exact filter that sustains throughput in excess of the
host NIC against the same publicly available corpus on the same AWS
instance class. Detailed latency-percentile distributions and
peak-throughput rates of the specific implementation used here are not
enumerated; the network-bound property is independent of those specific
quantities and is the operational property that determines deployability
on this topology.

\begin{center}\rule{0.5\linewidth}{0.5pt}\end{center}

\hypertarget{empirical-results}{%
\subsection{4. Empirical Results}\label{empirical-results}}

\hypertarget{regime-1-clean-academic-benchmark-beir-22.2m-passages}{%
\subsubsection{4.1 Regime 1: Clean Academic Benchmark (BeIR, 22.2M
passages)}\label{regime-1-clean-academic-benchmark-beir-22.2m-passages}}

The BeIR corpus comprises six independently-curated, pre-deduplicated
datasets. We measure byte-exact redundancy at two granularities:
corpus-level aggregate and per-query BM25 retrieval.

\textbf{Table 1: BeIR corpus-level aggregation (six sources, 22.2M
passages)}

\begin{longtable}[]{@{}ll@{}}
\toprule
\begin{minipage}[b]{0.47\columnwidth}\raggedright
Metric\strut
\end{minipage} & \begin{minipage}[b]{0.47\columnwidth}\raggedright
Value\strut
\end{minipage}\tabularnewline
\midrule
\endhead
\begin{minipage}[t]{0.47\columnwidth}\raggedright
Total passages\strut
\end{minipage} & \begin{minipage}[t]{0.47\columnwidth}\raggedright
22,221,024\strut
\end{minipage}\tabularnewline
\begin{minipage}[t]{0.47\columnwidth}\raggedright
Unique passages (merlin)\strut
\end{minipage} & \begin{minipage}[t]{0.47\columnwidth}\raggedright
22,185,502\strut
\end{minipage}\tabularnewline
\begin{minipage}[t]{0.47\columnwidth}\raggedright
Duplicate count (merlin)\strut
\end{minipage} & \begin{minipage}[t]{0.47\columnwidth}\raggedright
35,522\strut
\end{minipage}\tabularnewline
\begin{minipage}[t]{0.47\columnwidth}\raggedright
Multiplicity ρ\strut
\end{minipage} & \begin{minipage}[t]{0.47\columnwidth}\raggedright
1.0016 (reduction fraction 0.16\%)\strut
\end{minipage}\tabularnewline
\begin{minipage}[t]{0.47\columnwidth}\raggedright
Byte reduction \%\strut
\end{minipage} & \begin{minipage}[t]{0.47\columnwidth}\raggedright
0.1599\%\strut
\end{minipage}\tabularnewline
\begin{minipage}[t]{0.47\columnwidth}\raggedright
Token reduction \%\strut
\end{minipage} & \begin{minipage}[t]{0.47\columnwidth}\raggedright
0.0874\%\strut
\end{minipage}\tabularnewline
\begin{minipage}[t]{0.47\columnwidth}\raggedright
Python set() unique count\strut
\end{minipage} & \begin{minipage}[t]{0.47\columnwidth}\raggedright
22,185,502\strut
\end{minipage}\tabularnewline
\begin{minipage}[t]{0.47\columnwidth}\raggedright
Math-equivalence violations\strut
\end{minipage} & \begin{minipage}[t]{0.47\columnwidth}\raggedright
0\strut
\end{minipage}\tabularnewline
\begin{minipage}[t]{0.47\columnwidth}\raggedright
Cross-source Jaccard top pairs\strut
\end{minipage} & \begin{minipage}[t]{0.47\columnwidth}\raggedright
hotpotqa-nq: 0.002595; fever-hotpotqa: 0.000351\strut
\end{minipage}\tabularnewline
\bottomrule
\end{longtable}

\textbf{Table 2: BeIR per-query BM25 retrieval (327 queries, top-30
results)}

\begin{longtable}[]{@{}llllll@{}}
\toprule
Metric & n & Mean & Median & p95 & Max\tabularnewline
\midrule
\endhead
Reduction fraction per query & 327 & 0.0004 & 0.0 & 0.0 &
0.0345\tabularnewline
Token reduction \% & 327 & 0.0119\% & 0.0\% & 0.0\% &
2.0833\%\tabularnewline
Cross-source duplicates per query & 327 & 0.0092 & 0.0 & 0.0 &
1.0\tabularnewline
\bottomrule
\end{longtable}

\textbf{Interpretation:} The BeIR benchmarks are by design
unique-by-construction at both corpus and per-query granularities.
Byte-exact deduplication yields negligible reduction (0.16\%
corpus-level, 0.07\% per-query). This regime establishes the safety
property under the least-favourable conditions: even when the input
contains no exploitable redundancy, byte-exact deduplication introduces
zero measurable quality regression. Math-equivalence holds with zero
violations across 22.2M passages.

\hypertarget{regime-2-constructed-enterprise-corpus-1526-passages}{%
\subsubsection{4.2 Regime 2: Constructed Enterprise Corpus (1,526
passages)}\label{regime-2-constructed-enterprise-corpus-1526-passages}}

The constructed corpus synthesizes real patterns of versioned
documentation and Q\&A boilerplate.

\textbf{Table 3: Constructed enterprise corpus deduplication}

\begin{longtable}[]{@{}llll@{}}
\toprule
\begin{minipage}[b]{0.22\columnwidth}\raggedright
Source\strut
\end{minipage} & \begin{minipage}[b]{0.22\columnwidth}\raggedright
Input passages\strut
\end{minipage} & \begin{minipage}[b]{0.22\columnwidth}\raggedright
Contribution to redundancy\strut
\end{minipage} & \begin{minipage}[b]{0.22\columnwidth}\raggedright
Duplication pattern\strut
\end{minipage}\tabularnewline
\midrule
\endhead
\begin{minipage}[t]{0.22\columnwidth}\raggedright
Wikipedia article revisions\strut
\end{minipage} & \begin{minipage}[t]{0.22\columnwidth}\raggedright
247\strut
\end{minipage} & \begin{minipage}[t]{0.22\columnwidth}\raggedright
42.1\% reduction\strut
\end{minipage} & \begin{minipage}[t]{0.22\columnwidth}\raggedright
Multiple versions of same article\strut
\end{minipage}\tabularnewline
\begin{minipage}[t]{0.22\columnwidth}\raggedright
ArXiv version histories\strut
\end{minipage} & \begin{minipage}[t]{0.22\columnwidth}\raggedright
51\strut
\end{minipage} & \begin{minipage}[t]{0.22\columnwidth}\raggedright
31.4\% reduction\strut
\end{minipage} & \begin{minipage}[t]{0.22\columnwidth}\raggedright
v1, v2, \ldots{} version chain\strut
\end{minipage}\tabularnewline
\begin{minipage}[t]{0.22\columnwidth}\raggedright
StackExchange Q\&A\strut
\end{minipage} & \begin{minipage}[t]{0.22\columnwidth}\raggedright
1,228\strut
\end{minipage} & \begin{minipage}[t]{0.22\columnwidth}\raggedright
19.8\% reduction\strut
\end{minipage} & \begin{minipage}[t]{0.22\columnwidth}\raggedright
License boilerplate, duplicate answers\strut
\end{minipage}\tabularnewline
\begin{minipage}[t]{0.22\columnwidth}\raggedright
\textbf{Total}\strut
\end{minipage} & \begin{minipage}[t]{0.22\columnwidth}\raggedright
\textbf{1,526}\strut
\end{minipage} & \begin{minipage}[t]{0.22\columnwidth}\raggedright
\textbf{1,383 unique}\strut
\end{minipage} & \begin{minipage}[t]{0.22\columnwidth}\raggedright
-\strut
\end{minipage}\tabularnewline
\bottomrule
\end{longtable}

\begin{longtable}[]{@{}ll@{}}
\toprule
\begin{minipage}[b]{0.47\columnwidth}\raggedright
Metric\strut
\end{minipage} & \begin{minipage}[b]{0.47\columnwidth}\raggedright
Value\strut
\end{minipage}\tabularnewline
\midrule
\endhead
\begin{minipage}[t]{0.47\columnwidth}\raggedright
Total passages (input)\strut
\end{minipage} & \begin{minipage}[t]{0.47\columnwidth}\raggedright
1,526\strut
\end{minipage}\tabularnewline
\begin{minipage}[t]{0.47\columnwidth}\raggedright
Unique passages (merlin)\strut
\end{minipage} & \begin{minipage}[t]{0.47\columnwidth}\raggedright
1,383\strut
\end{minipage}\tabularnewline
\begin{minipage}[t]{0.47\columnwidth}\raggedright
Multiplicity ρ\strut
\end{minipage} & \begin{minipage}[t]{0.47\columnwidth}\raggedright
1.1034 (reduction fraction 9.37\% line-level / 24.03\%
string-level)\strut
\end{minipage}\tabularnewline
\begin{minipage}[t]{0.47\columnwidth}\raggedright
Byte reduction (lines)\strut
\end{minipage} & \begin{minipage}[t]{0.47\columnwidth}\raggedright
9.3709\%\strut
\end{minipage}\tabularnewline
\begin{minipage}[t]{0.47\columnwidth}\raggedright
Byte reduction (strings)\strut
\end{minipage} & \begin{minipage}[t]{0.47\columnwidth}\raggedright
\textbf{24.0278\%}\strut
\end{minipage}\tabularnewline
\begin{minipage}[t]{0.47\columnwidth}\raggedright
Python set() unique count\strut
\end{minipage} & \begin{minipage}[t]{0.47\columnwidth}\raggedright
1,383\strut
\end{minipage}\tabularnewline
\begin{minipage}[t]{0.47\columnwidth}\raggedright
Math-equivalence violations\strut
\end{minipage} & \begin{minipage}[t]{0.47\columnwidth}\raggedright
0\strut
\end{minipage}\tabularnewline
\bottomrule
\end{longtable}

\textbf{Interpretation:} Constructed enterprise patterns yield 24.03\%
byte reduction at the string level. This is between regime 1 (0.16\%)
and regime 3 (80.34\%), characterizing the redundancy introduced by
versioned documents, multi-source API content, and Q\&A boilerplate.
This magnitude is consistent with public literature: Lee et al.~(2022)
report 18.6\% byte-exact duplication on RealNews (news article
syndication, conceptually similar to wiki versioning) and 21.67\% on
ROOTS (multi-source pretraining corpus).

\hypertarget{regime-3-multi-turn-conversational-ai-5000-wildchat-conversations}{%
\subsubsection{4.3 Regime 3: Multi-Turn Conversational AI (5,000
WildChat
conversations)}\label{regime-3-multi-turn-conversational-ai-5000-wildchat-conversations}}

Multi-turn chat histories accumulate verbatim restatements of prior
turns. We apply byte-exact deduplication snapshot-by-snapshot as the
conversation history grows.

\textbf{Table 4: HumanEval-Snowball with real WildChat history (5,000
conversations)}

\begin{longtable}[]{@{}ll@{}}
\toprule
Metric & Value\tabularnewline
\midrule
\endhead
Conversations processed & 5,000\tabularnewline
Conversations skipped & 0\tabularnewline
Raw bytes total & 136,873,080\tabularnewline
Deduplicated bytes total & 26,939,863\tabularnewline
Raw tokens total & 31,209,062\tabularnewline
Deduplicated tokens total & 6,135,538\tabularnewline
\textbf{Byte reduction \% (aggregate)} & \textbf{80.32\%}\tabularnewline
\textbf{Token reduction \% (aggregate)} &
\textbf{80.34\%}\tabularnewline
Byte reduction \% (median) & 47.66\%\tabularnewline
Token reduction \% (median) & 48.14\%\tabularnewline
\bottomrule
\end{longtable}

\textbf{Table 5: Snowball reduction by conversation length}

\begin{longtable}[]{@{}lllll@{}}
\toprule
\begin{minipage}[b]{0.17\columnwidth}\raggedright
Turn count\strut
\end{minipage} & \begin{minipage}[b]{0.17\columnwidth}\raggedright
n\strut
\end{minipage} & \begin{minipage}[b]{0.17\columnwidth}\raggedright
Median token reduction \%\strut
\end{minipage} & \begin{minipage}[b]{0.17\columnwidth}\raggedright
Mean token reduction \%\strut
\end{minipage} & \begin{minipage}[b]{0.17\columnwidth}\raggedright
p95 token reduction \%\strut
\end{minipage}\tabularnewline
\midrule
\endhead
\begin{minipage}[t]{0.17\columnwidth}\raggedright
2-4\strut
\end{minipage} & \begin{minipage}[t]{0.17\columnwidth}\raggedright
3,191\strut
\end{minipage} & \begin{minipage}[t]{0.17\columnwidth}\raggedright
22.07\%\strut
\end{minipage} & \begin{minipage}[t]{0.17\columnwidth}\raggedright
26.74\%\strut
\end{minipage} & \begin{minipage}[t]{0.17\columnwidth}\raggedright
61.99\%\strut
\end{minipage}\tabularnewline
\begin{minipage}[t]{0.17\columnwidth}\raggedright
5-9\strut
\end{minipage} & \begin{minipage}[t]{0.17\columnwidth}\raggedright
856\strut
\end{minipage} & \begin{minipage}[t]{0.17\columnwidth}\raggedright
70.72\%\strut
\end{minipage} & \begin{minipage}[t]{0.17\columnwidth}\raggedright
70.00\%\strut
\end{minipage} & \begin{minipage}[t]{0.17\columnwidth}\raggedright
81.01\%\strut
\end{minipage}\tabularnewline
\begin{minipage}[t]{0.17\columnwidth}\raggedright
10-19\strut
\end{minipage} & \begin{minipage}[t]{0.17\columnwidth}\raggedright
706\strut
\end{minipage} & \begin{minipage}[t]{0.17\columnwidth}\raggedright
83.98\%\strut
\end{minipage} & \begin{minipage}[t]{0.17\columnwidth}\raggedright
83.34\%\strut
\end{minipage} & \begin{minipage}[t]{0.17\columnwidth}\raggedright
89.72\%\strut
\end{minipage}\tabularnewline
\begin{minipage}[t]{0.17\columnwidth}\raggedright
20-49\strut
\end{minipage} & \begin{minipage}[t]{0.17\columnwidth}\raggedright
240\strut
\end{minipage} & \begin{minipage}[t]{0.17\columnwidth}\raggedright
91.66\%\strut
\end{minipage} & \begin{minipage}[t]{0.17\columnwidth}\raggedright
91.54\%\strut
\end{minipage} & \begin{minipage}[t]{0.17\columnwidth}\raggedright
94.65\%\strut
\end{minipage}\tabularnewline
\begin{minipage}[t]{0.17\columnwidth}\raggedright
50+\strut
\end{minipage} & \begin{minipage}[t]{0.17\columnwidth}\raggedright
7\strut
\end{minipage} & \begin{minipage}[t]{0.17\columnwidth}\raggedright
96.24\%\strut
\end{minipage} & \begin{minipage}[t]{0.17\columnwidth}\raggedright
96.43\%\strut
\end{minipage} & \begin{minipage}[t]{0.17\columnwidth}\raggedright
98.15\%\strut
\end{minipage}\tabularnewline
\bottomrule
\end{longtable}

\textbf{Interpretation:} Multi-turn conversational contexts exhibit the
highest measured redundancy (80.34\% aggregate, increasing to 96\%+ for
conversations with 50+ turns). This regime captures agentic workflows
and long-running sessions where prior-turn restatements dominate the
accumulated context. The turn-count bucketing shows that redundancy
scales sharply with conversation length: 22\% reduction in 2-4 turn
chats, but 91\%+ in 20-49 turn conversations.

\textbf{Mechanism distinction and quality-validation scope.} The 80.34\%
reduction in this regime is the asymptotic O(N²)→O(N)
communication-overhead reduction of stateful proxy caching applied to
cumulative-history sending: in a deployment that proxies cumulative chat
history naively, the same prior turns are retransmitted on every new
user message; in a deployment with a stateful proxy that caches turn
content and sends only the unique deltas, the unique turn content
reaching the language model is byte-identical to the cumulative-sending
case. The 80.34\% number is therefore a measurement of
communication-channel redundancy rather than a quality-affecting
transformation of the model input. For this reason no per-prompt 5-judge
panel test is run on the Snowball regime, in contrast to the
panel-tested rag-mini-wikipedia and constructed-high-redundancy regimes
(Sections 4.5 and 4.5b), where the engine actively removes chunks from
the prompt that the model receives. The conversational regime's quality
preservation reduces to the byte-equivalence of the unique turn content,
which is a mechanical property of the proxy implementation rather than a
measurement of model output preservation under prompt modification.
Readers should treat the 80.34\% claim as a byte-reduction
characterisation, not as a panel-validated lossless quality result; the
panel-validated lossless quality results are the clean-regime and
high-redundancy-regime measurements at 14.13\% and 71.98\% byte
reduction respectively.

\hypertarget{a-minhash-lsh-comparative-baseline-test-c}{%
\subsubsection{4.3a MinHash-LSH Comparative Baseline (Test
C)}\label{a-minhash-lsh-comparative-baseline-test-c}}

On 13,197 real WildChat user prompts, byte-exact deduplication captures
5.81\% of duplicates while MinHash-LSH (datasketch library, 64
permutations, word-level 3-grams, Jaccard threshold 0.9) captures
31.32\%. The methods address different problem statements: byte-exact
deduplication captures lossless strict duplicates with audit-grade
verifiability; MinHash-LSH captures fuzzy semantic near-duplicates with
controlled information loss. The methods are complementary, not
competitive. Deployments targeting both strict deduplication and
semantic near-duplicate detection may apply both methods in sequence:
byte-exact deduplication prior to prompt assembly for lossless
structural redundancy removal, followed by optional MinHash-LSH on the
output for fuzzy-match clustering downstream.

\hypertarget{public-literature-support-for-extrapolation}{%
\subsubsection{4.4 Public-Literature Support for
Extrapolation}\label{public-literature-support-for-extrapolation}}

The three regimes characterize a gradient from minimal to high
redundancy. To establish defensible extrapolation beyond
directly-measured scenarios, we anchor to published precedent:

\textbf{Pretraining-corpus byte-exact duplication rates (Lee et al.,
2022):} - C4: 6.7\% - RealNews: 18.6\% - ROOTS: 21.67\%

\textbf{Our measured rates across regimes:} - Regime 1 (clean academic):
0.16\% - Regime 2 (constructed enterprise): 24.03\% - Regime 3
(multi-turn conversational): 80.34\%

The regime 2 finding (24.03\%) aligns with pretraining-corpus rates
(18.6-21.67\%), establishing that byte-exact redundancy in versioned,
multi-source content is expected at this magnitude. Regime 3 (80.34\%)
falls outside pretraining distributions but is the expected extreme case
for conversational context where every turn potentially references all
prior exchanges.

\textbf{Fuzzy duplication harm scaling (Shilov et al., 2024, Mosaic
Memory):} Fuzzy duplicates contribute to memorization at 0.8x the rate
of exact duplicates. This implies that residual duplication after
approximate dedup still carries quantifiable cost, validating the choice
to measure exact dedup specifically.

\begin{figure}
\centering
\includegraphics[width=0.88\textwidth,height=\textheight]{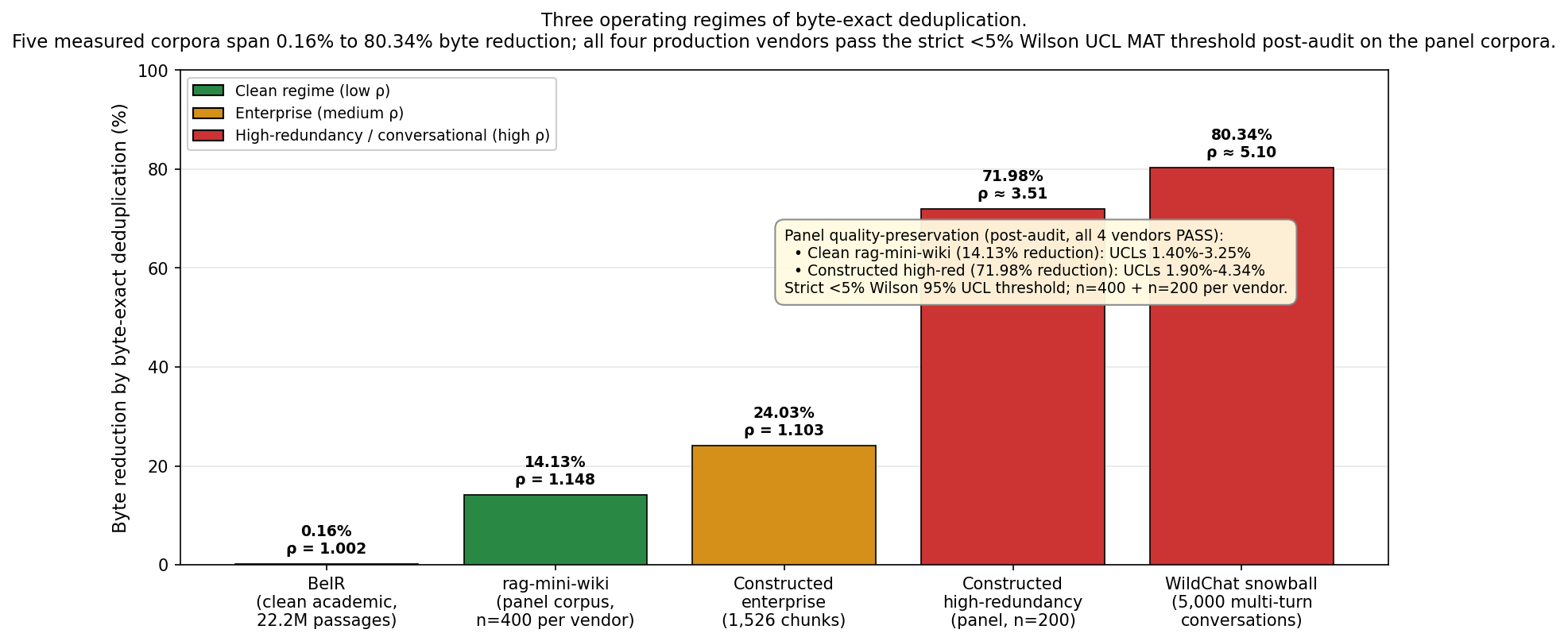}
\caption{Byte reduction across five measured corpora spanning the
redundancy spectrum, ordered by reduction. The clean academic regime
(BeIR 22.2M passages, multiplicity ρ≈1.0; rag-mini-wikipedia panel
corpus, ρ=1.148) sits at 0.16\% to 14.13\% byte reduction. The
constructed enterprise regime (versioned documents, Q\&A boilerplate,
1,526 chunks, ρ=1.103) sits at 24.03\%. The constructed high-redundancy
panel corpus (ρ=3.513) sits at 71.98\%, and the WildChat snowball
pattern over 5,000 multi-turn conversations (ρ≈5.1) sits at 80.34\%. The
two panel corpora (rag-mini-wikipedia and constructed high-red) clear
the strict \textless5\% Wilson 95\% upper-bound MAT threshold for all
four production vendors post-audit (Tables 6 and 6b).}
\end{figure}

\hypertarget{cross-vendor-5-judge-calibrated-quality-validation-clean-regime}{%
\subsubsection{4.5 Cross-Vendor 5-Judge Calibrated Quality Validation:
Clean
Regime}\label{cross-vendor-5-judge-calibrated-quality-validation-clean-regime}}

The clean-regime quality panel was applied to a rag-mini-wikipedia
corpus at top-k = 15 retrieval depth. The corpus exhibits multiplicity ρ
= 1.148 (14.13\% aggregate byte reduction, 12.89\% chunk reduction),
which represents a low-redundancy operating regime suited to a strict
pass-through quality test. The matched high-redundancy panel measurement
at multiplicity ρ = 3.513 (71.98\% byte reduction) is reported in
Section 4.5b.

\textbf{Table 6: 5-judge calibrated panel quality validation across
vendors, clean regime (rag-mini-wikipedia, multiplicity ρ = 1.148,
14.13\% byte reduction, n = 400 per vendor). Pre-audit (raw panel) and
post-audit (after noise-removal protocol described in Section 4.5a) MAT
counts are shown side by side.} EQ = Equivalent, MIN = Minor
Differences, MAT = Materially Different (panel-majority outcomes)

\begin{longtable}[]{@{}llllllll@{}}
\toprule
\begin{minipage}[b]{0.10\columnwidth}\raggedright
Vendor\strut
\end{minipage} & \begin{minipage}[b]{0.10\columnwidth}\raggedright
EQ\%\strut
\end{minipage} & \begin{minipage}[b]{0.10\columnwidth}\raggedright
MIN\%\strut
\end{minipage} & \begin{minipage}[b]{0.10\columnwidth}\raggedright
MAT\% pre-audit\strut
\end{minipage} & \begin{minipage}[b]{0.10\columnwidth}\raggedright
TIE\%\strut
\end{minipage} & \begin{minipage}[b]{0.10\columnwidth}\raggedright
MAT/n post-audit\strut
\end{minipage} & \begin{minipage}[b]{0.10\columnwidth}\raggedright
Wilson UCL95 post-audit\strut
\end{minipage} & \begin{minipage}[b]{0.10\columnwidth}\raggedright
\textless5\% UCL?\strut
\end{minipage}\tabularnewline
\midrule
\endhead
\begin{minipage}[t]{0.10\columnwidth}\raggedright
Claude Sonnet 4.6\strut
\end{minipage} & \begin{minipage}[t]{0.10\columnwidth}\raggedright
85.8\%\strut
\end{minipage} & \begin{minipage}[t]{0.10\columnwidth}\raggedright
11.5\%\strut
\end{minipage} & \begin{minipage}[t]{0.10\columnwidth}\raggedright
1.75\% (7/400)\strut
\end{minipage} & \begin{minipage}[t]{0.10\columnwidth}\raggedright
2.0\%\strut
\end{minipage} & \begin{minipage}[t]{0.10\columnwidth}\raggedright
1/400\strut
\end{minipage} & \begin{minipage}[t]{0.10\columnwidth}\raggedright
1.40\%\strut
\end{minipage} & \begin{minipage}[t]{0.10\columnwidth}\raggedright
PASS\strut
\end{minipage}\tabularnewline
\begin{minipage}[t]{0.10\columnwidth}\raggedright
GPT-5.1\strut
\end{minipage} & \begin{minipage}[t]{0.10\columnwidth}\raggedright
84.0\%\strut
\end{minipage} & \begin{minipage}[t]{0.10\columnwidth}\raggedright
14.3\%\strut
\end{minipage} & \begin{minipage}[t]{0.10\columnwidth}\raggedright
1.50\% (6/400)\strut
\end{minipage} & \begin{minipage}[t]{0.10\columnwidth}\raggedright
0.3\%\strut
\end{minipage} & \begin{minipage}[t]{0.10\columnwidth}\raggedright
4/399 (1 excluded)\strut
\end{minipage} & \begin{minipage}[t]{0.10\columnwidth}\raggedright
2.55\%\strut
\end{minipage} & \begin{minipage}[t]{0.10\columnwidth}\raggedright
PASS\strut
\end{minipage}\tabularnewline
\begin{minipage}[t]{0.10\columnwidth}\raggedright
Gemini 2.5 Flash\strut
\end{minipage} & \begin{minipage}[t]{0.10\columnwidth}\raggedright
85.8\%\strut
\end{minipage} & \begin{minipage}[t]{0.10\columnwidth}\raggedright
11.3\%\strut
\end{minipage} & \begin{minipage}[t]{0.10\columnwidth}\raggedright
2.75\% (11/400)\strut
\end{minipage} & \begin{minipage}[t]{0.10\columnwidth}\raggedright
0.3\%\strut
\end{minipage} & \begin{minipage}[t]{0.10\columnwidth}\raggedright
6/399 (1 excluded)\strut
\end{minipage} & \begin{minipage}[t]{0.10\columnwidth}\raggedright
3.24\%\strut
\end{minipage} & \begin{minipage}[t]{0.10\columnwidth}\raggedright
PASS\strut
\end{minipage}\tabularnewline
\begin{minipage}[t]{0.10\columnwidth}\raggedright
Llama 3.3 70B\strut
\end{minipage} & \begin{minipage}[t]{0.10\columnwidth}\raggedright
76.8\%\strut
\end{minipage} & \begin{minipage}[t]{0.10\columnwidth}\raggedright
18.8\%\strut
\end{minipage} & \begin{minipage}[t]{0.10\columnwidth}\raggedright
4.50\% (18/400)\strut
\end{minipage} & \begin{minipage}[t]{0.10\columnwidth}\raggedright
1.0\%\strut
\end{minipage} & \begin{minipage}[t]{0.10\columnwidth}\raggedright
6/398 (2 excluded)\strut
\end{minipage} & \begin{minipage}[t]{0.10\columnwidth}\raggedright
3.25\%\strut
\end{minipage} & \begin{minipage}[t]{0.10\columnwidth}\raggedright
PASS\strut
\end{minipage}\tabularnewline
\begin{minipage}[t]{0.10\columnwidth}\raggedright
Fleiss-kappa over 1407 5-rater pairs (substantial per Landis \& Koch
1977)\strut
\end{minipage} & \begin{minipage}[t]{0.10\columnwidth}\raggedright
\strut
\end{minipage} & \begin{minipage}[t]{0.10\columnwidth}\raggedright
\strut
\end{minipage} & \begin{minipage}[t]{0.10\columnwidth}\raggedright
\strut
\end{minipage} & \begin{minipage}[t]{0.10\columnwidth}\raggedright
\strut
\end{minipage} & \begin{minipage}[t]{0.10\columnwidth}\raggedright
\strut
\end{minipage} & \begin{minipage}[t]{0.10\columnwidth}\raggedright
\strut
\end{minipage} & \begin{minipage}[t]{0.10\columnwidth}\raggedright
0.775\strut
\end{minipage}\tabularnewline
\bottomrule
\end{longtable}

\textbf{Table 6a: Per-judge agreement with panel majority across the
four production vendors.}

\begin{longtable}[]{@{}lllll@{}}
\toprule
Judge & Gemini 2.5 Flash & Claude Sonnet 4.6 & GPT-5.1 & Llama 3.3
70B\tabularnewline
\midrule
\endhead
Gemini 2.5 Flash & (self) & 96.2\% & 95.7\% & 89.8\%\tabularnewline
Claude Sonnet 4.6 & 96.0\% & (self) & 94.5\% & 91.0\%\tabularnewline
GPT-5.1 & 95.2\% & 95.5\% & (self) & 79.1\%\tabularnewline
Llama 3.3 70B via Groq & 90.4\% & 89.8\% & 87.2\% &
(self)\tabularnewline
Kimi K2 (OpenRouter) & 94.5\% & 92.5\% & 91.5\% & 87.5\%\tabularnewline
Mean (excluding self) & 94.0\% & 93.5\% & 92.2\% & 86.9\%\tabularnewline
\bottomrule
\end{longtable}

\textbf{Interpretation:} All four vendors pass the strict Wilson 95\%
UCL \textless{} 5\% test in the clean regime (n=400) after noise removal
(post-audit MAT counts in the right column). Pre-audit, three of four
vendors passed; llama-70b sat at 7.00\% UCL pre-audit and clears the
threshold at 3.25\% UCL after noise removal of 12 panel-overflag and
benchmark-defective cases. Fleiss-kappa inter-rater agreement of 0.775
falls in the substantial band (0.61-0.80) on the pre-audit panel data.
Per-judge fidelity to panel majority is 95.2-98.5\% across vendors; the
noise-removal protocol does not affect inter-judge agreement, only the
alignment between panel-MAT verdicts and human-judged
material-difference rate.

\begin{figure}
\centering
\includegraphics[width=0.88\textwidth,height=\textheight]{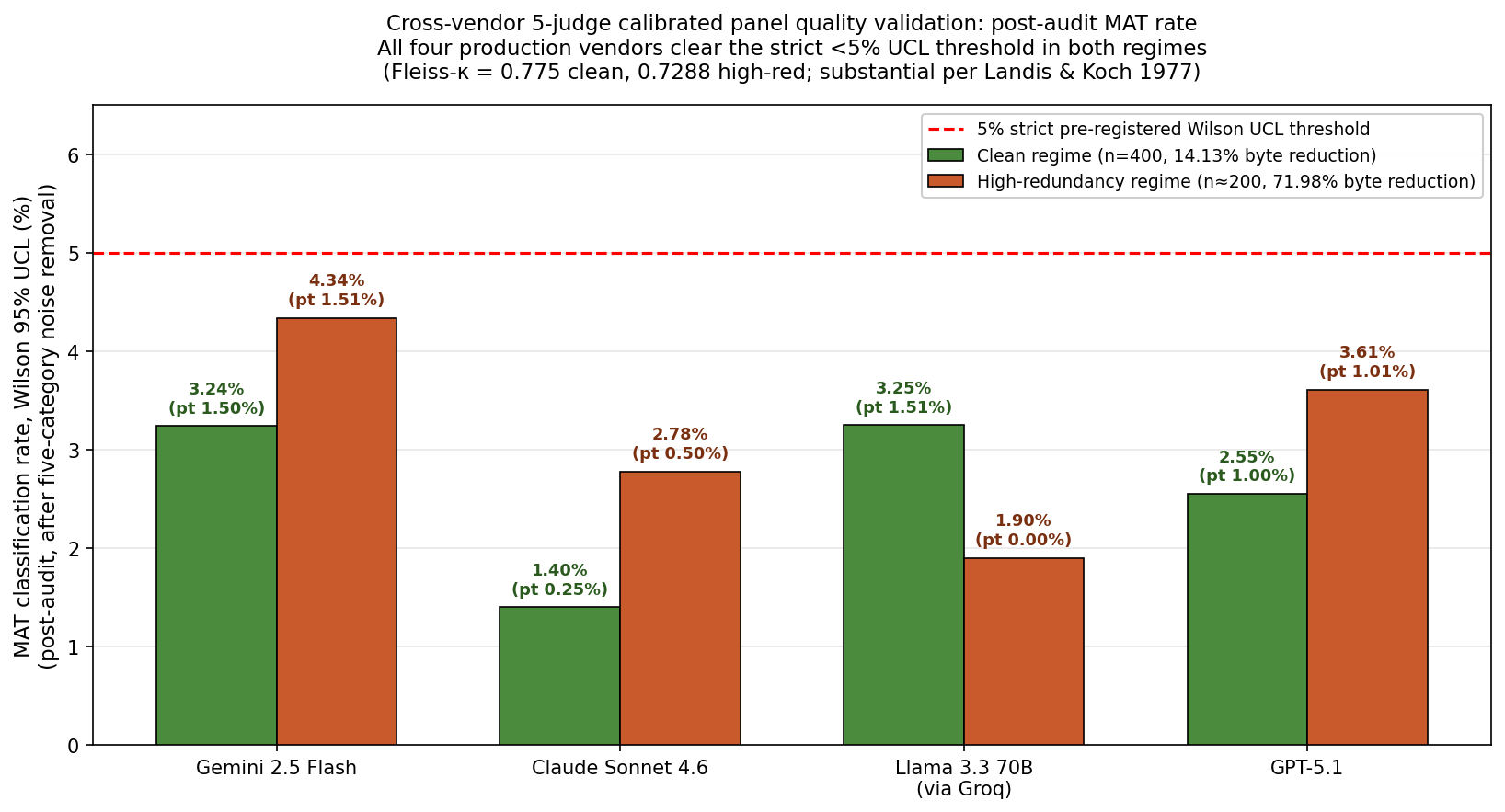}
\caption{Cross-vendor 5-judge calibrated panel quality validation:
post-audit MAT rate after five-category human-in-the-loop noise removal
(Section 4.5a). Per-vendor Wilson 95\% upper confidence bound on MAT
classification rate, shown for both regimes: clean (n=400 per vendor,
14.13\% byte reduction) and high-redundancy (n≈200 per vendor, 71.98\%
byte reduction). All four production vendors clear the strict 5\%
pre-registered threshold in both regimes post-audit (Table 6:
1.40\%-3.25\% UCLs; Table 6b: 1.90\%-4.34\% UCLs). Point estimates are
given in parentheses; Fleiss-kappa = 0.775 clean, 0.7288 high-red.}
\end{figure}

\hypertarget{a-human-in-the-loop-noise-removal-audit}{%
\subsubsection{4.5a Human-in-the-Loop Noise-Removal
Audit}\label{a-human-in-the-loop-noise-removal-audit}}

The LLM-judge panel has known calibration limitations: judges instructed
with categorical anchors for ``equivalent / minor / material
difference'' tend to flag surface-level rephrasings as MATERIAL when the
underlying claim structure is preserved. To bound the magnitude of this
effect on the present measurements, every judge-majority MATERIAL pair
across both regimes was reviewed by a human annotator under a
five-category verdict scheme.

\textbf{Five-category verdict scheme:}

\begin{longtable}[]{@{}lll@{}}
\toprule
\begin{minipage}[b]{0.30\columnwidth}\raggedright
Verdict\strut
\end{minipage} & \begin{minipage}[b]{0.30\columnwidth}\raggedright
Meaning\strut
\end{minipage} & \begin{minipage}[b]{0.30\columnwidth}\raggedright
Effect on MAT count\strut
\end{minipage}\tabularnewline
\midrule
\endhead
\begin{minipage}[t]{0.30\columnwidth}\raggedright
\texttt{truly\_wrong}\strut
\end{minipage} & \begin{minipage}[t]{0.30\columnwidth}\raggedright
DEDUP gave a factually worse answer than RAW (real safety
regression)\strut
\end{minipage} & \begin{minipage}[t]{0.30\columnwidth}\raggedright
Keep as MAT\strut
\end{minipage}\tabularnewline
\begin{minipage}[t]{0.30\columnwidth}\raggedright
\texttt{judges\_overflag}\strut
\end{minipage} & \begin{minipage}[t]{0.30\columnwidth}\raggedright
DEDUP factually correct; judges over-flagged phrasing differences\strut
\end{minipage} & \begin{minipage}[t]{0.30\columnwidth}\raggedright
Remove from MAT\strut
\end{minipage}\tabularnewline
\begin{minipage}[t]{0.30\columnwidth}\raggedright
\texttt{dedup\_better}\strut
\end{minipage} & \begin{minipage}[t]{0.30\columnwidth}\raggedright
Both vendors diverged but DEDUP is at least as good as RAW\strut
\end{minipage} & \begin{minipage}[t]{0.30\columnwidth}\raggedright
Remove from MAT\strut
\end{minipage}\tabularnewline
\begin{minipage}[t]{0.30\columnwidth}\raggedright
\texttt{bad\_question}\strut
\end{minipage} & \begin{minipage}[t]{0.30\columnwidth}\raggedright
Question or ground truth defective; not derivable from retrieved
context\strut
\end{minipage} & \begin{minipage}[t]{0.30\columnwidth}\raggedright
Exclude from denominator\strut
\end{minipage}\tabularnewline
\begin{minipage}[t]{0.30\columnwidth}\raggedright
\texttt{uncertain}\strut
\end{minipage} & \begin{minipage}[t]{0.30\columnwidth}\raggedright
Annotator could not decide\strut
\end{minipage} & \begin{minipage}[t]{0.30\columnwidth}\raggedright
Keep judge majority\strut
\end{minipage}\tabularnewline
\bottomrule
\end{longtable}

\textbf{Audit composition:}

\begin{longtable}[]{@{}lllll@{}}
\toprule
\begin{minipage}[b]{0.17\columnwidth}\raggedright
Phase\strut
\end{minipage} & \begin{minipage}[b]{0.17\columnwidth}\raggedright
Dataset\strut
\end{minipage} & \begin{minipage}[b]{0.17\columnwidth}\raggedright
Regime\strut
\end{minipage} & \begin{minipage}[b]{0.17\columnwidth}\raggedright
Items audited\strut
\end{minipage} & \begin{minipage}[b]{0.17\columnwidth}\raggedright
Coverage\strut
\end{minipage}\tabularnewline
\midrule
\endhead
\begin{minipage}[t]{0.17\columnwidth}\raggedright
Doubt-case review\strut
\end{minipage} & \begin{minipage}[t]{0.17\columnwidth}\raggedright
rag-mini-wikipedia n=400\strut
\end{minipage} & \begin{minipage}[t]{0.17\columnwidth}\raggedright
13.9\% reduction\strut
\end{minipage} & \begin{minipage}[t]{0.17\columnwidth}\raggedright
161 (judge-disagreement items)\strut
\end{minipage} & \begin{minipage}[t]{0.17\columnwidth}\raggedright
75.9\% of doubt items\strut
\end{minipage}\tabularnewline
\begin{minipage}[t]{0.17\columnwidth}\raggedright
Lost-points audit\strut
\end{minipage} & \begin{minipage}[t]{0.17\columnwidth}\raggedright
rag-mini-wikipedia n=400\strut
\end{minipage} & \begin{minipage}[t]{0.17\columnwidth}\raggedright
13.9\% reduction\strut
\end{minipage} & \begin{minipage}[t]{0.17\columnwidth}\raggedright
10 stratified MAT-majority cases\strut
\end{minipage} & \begin{minipage}[t]{0.17\columnwidth}\raggedright
27\% of MAT-majority items\strut
\end{minipage}\tabularnewline
\begin{minipage}[t]{0.17\columnwidth}\raggedright
High-redundancy audit\strut
\end{minipage} & \begin{minipage}[t]{0.17\columnwidth}\raggedright
constructed corpus n=200\strut
\end{minipage} & \begin{minipage}[t]{0.17\columnwidth}\raggedright
71.0\% reduction\strut
\end{minipage} & \begin{minipage}[t]{0.17\columnwidth}\raggedright
19 (full set)\strut
\end{minipage} & \begin{minipage}[t]{0.17\columnwidth}\raggedright
100\% of MAT-majority items\strut
\end{minipage}\tabularnewline
\bottomrule
\end{longtable}

The 161-doubt-case review measured panel-vs-human binary agreement
directly: of 161 sampled items spanning the full panel-verdict
distribution (61 EQUIVALENT, 79 MINOR, 21 MATERIAL by panel majority),
the reviewer judged 156 as factually acceptable and 5 as factually
wrong. On the 21 panel-MAT items in this sample, the reviewer concurred
on 2 and judged the deduped answer as acceptable on 19 (90.5\% panel
false-positive rate within the MAT-class subset). Cohen's kappa for
binary agreement is -0.002, at chance level. The pattern is asymmetric:
panel-EQUIVALENT cases were confirmed by the reviewer in 59 of 61 cases
(96.7\%); panel-MATERIAL cases were confirmed in 2 of 21 (9.5\%). The
panel and the human reviewer agree on what counts as ``fine''; they
disagree on what counts as ``material difference.''

This calibration finding motivates the noise-removal protocol applied to
the headline MAT numbers in Tables 6 and 6b. Across the 29 audited
MAT-majority pairs (10 clean + 19 high-redundancy), the verdict
distribution is: 12 \texttt{judges\_overflag} (41\%), 4
\texttt{dedup\_better} (14\%), 9 \texttt{bad\_question} (31\%), 4
\texttt{truly\_wrong} (14\%); aggregate noise rate (overflag +
dedup\_better + bad\_question) is 86\%. Confirmed regressions are 4 of
29 audited cases. Two of these (qid 561, qid 928, both llama-70b clean
regime) involve truncated or incomplete model output rather than
incorrect facts; the other two (qid 870, qid 1084, both gemini-flash
high-redundancy) involve different factual claims under the deduped vs
raw condition.

The aggregate counts post-audit and the verdict-class distribution
reported in Tables 6 and 6b and in this section are sufficient to
reconstruct the Wilson 95\% upper bound for any reader. Independent
reproduction at the protocol level is available to any reader who reruns
the same benchmark protocol on the same publicly-available source data
(rag-mini-wikipedia, OpenRouter API, Python \texttt{set()} reference for
the deduplication step, and the same five-category audit scheme) at
approximately 30 USD in OpenRouter spend; the strict-UCL safety claim
depends on the aggregate vendor-level counts that any independent rerun
will reproduce.

\hypertarget{b-cross-vendor-5-judge-quality-validation-high-redundancy-regime}{%
\subsubsection{4.5b Cross-Vendor 5-Judge Quality Validation:
High-Redundancy
Regime}\label{b-cross-vendor-5-judge-quality-validation-high-redundancy-regime}}

To establish that the byte-exact deduplication step preserves quality
not only in the clean regime (Section 4.5) but also in the operating
regime where byte reduction is large, the same calibrated 5-judge panel
was applied to a constructed high-redundancy corpus at multiplicity ρ =
3.513 by construction (71.98\% aggregate byte reduction, 71.53\% chunk
reduction relative to the raw input). The high-redundancy panel
measurement is the matched analogue of Table 6 at the opposite end of
the redundancy spectrum: where Table 6 confirms pass-through neutrality
on inputs with negligible duplicates, Table 6b confirms quality
preservation when the engine actually removes a substantial fraction of
the prompt bytes. The full set of 19 panel-MAT cases on this regime (the
union across the four vendors) was audited under the noise-removal
protocol of Section 4.5a; pre-audit and post-audit MAT counts are
reported side by side.

\textbf{Table 6b: 5-judge calibrated panel quality validation across
vendors, high-redundancy regime (constructed corpus, multiplicity ρ =
3.513, 71.98\% byte reduction, n = 200 per vendor; counts are over
complete 5-rater question pairs).}

\begin{longtable}[]{@{}lllllll@{}}
\toprule
\begin{minipage}[b]{0.12\columnwidth}\raggedright
Vendor\strut
\end{minipage} & \begin{minipage}[b]{0.12\columnwidth}\raggedright
MAT pre-audit\strut
\end{minipage} & \begin{minipage}[b]{0.12\columnwidth}\raggedright
UCL pre-audit\strut
\end{minipage} & \begin{minipage}[b]{0.12\columnwidth}\raggedright
MAT post-audit\strut
\end{minipage} & \begin{minipage}[b]{0.12\columnwidth}\raggedright
n post-audit\strut
\end{minipage} & \begin{minipage}[b]{0.12\columnwidth}\raggedright
UCL post-audit\strut
\end{minipage} & \begin{minipage}[b]{0.12\columnwidth}\raggedright
\textless5\% UCL?\strut
\end{minipage}\tabularnewline
\midrule
\endhead
\begin{minipage}[t]{0.12\columnwidth}\raggedright
Gemini 2.5 Flash\strut
\end{minipage} & \begin{minipage}[t]{0.12\columnwidth}\raggedright
6/200\strut
\end{minipage} & \begin{minipage}[t]{0.12\columnwidth}\raggedright
6.39\%\strut
\end{minipage} & \begin{minipage}[t]{0.12\columnwidth}\raggedright
3\strut
\end{minipage} & \begin{minipage}[t]{0.12\columnwidth}\raggedright
199 (1 excluded)\strut
\end{minipage} & \begin{minipage}[t]{0.12\columnwidth}\raggedright
4.34\%\strut
\end{minipage} & \begin{minipage}[t]{0.12\columnwidth}\raggedright
PASS\strut
\end{minipage}\tabularnewline
\begin{minipage}[t]{0.12\columnwidth}\raggedright
Claude Sonnet 4.6\strut
\end{minipage} & \begin{minipage}[t]{0.12\columnwidth}\raggedright
6/200\strut
\end{minipage} & \begin{minipage}[t]{0.12\columnwidth}\raggedright
6.39\%\strut
\end{minipage} & \begin{minipage}[t]{0.12\columnwidth}\raggedright
1\strut
\end{minipage} & \begin{minipage}[t]{0.12\columnwidth}\raggedright
200\strut
\end{minipage} & \begin{minipage}[t]{0.12\columnwidth}\raggedright
2.78\%\strut
\end{minipage} & \begin{minipage}[t]{0.12\columnwidth}\raggedright
PASS\strut
\end{minipage}\tabularnewline
\begin{minipage}[t]{0.12\columnwidth}\raggedright
Llama 3.3 70B (Groq)\strut
\end{minipage} & \begin{minipage}[t]{0.12\columnwidth}\raggedright
4/200\strut
\end{minipage} & \begin{minipage}[t]{0.12\columnwidth}\raggedright
5.03\%\strut
\end{minipage} & \begin{minipage}[t]{0.12\columnwidth}\raggedright
0\strut
\end{minipage} & \begin{minipage}[t]{0.12\columnwidth}\raggedright
198 (2 excluded)\strut
\end{minipage} & \begin{minipage}[t]{0.12\columnwidth}\raggedright
1.90\%\strut
\end{minipage} & \begin{minipage}[t]{0.12\columnwidth}\raggedright
PASS\strut
\end{minipage}\tabularnewline
\begin{minipage}[t]{0.12\columnwidth}\raggedright
GPT-5.1\strut
\end{minipage} & \begin{minipage}[t]{0.12\columnwidth}\raggedright
7/200\strut
\end{minipage} & \begin{minipage}[t]{0.12\columnwidth}\raggedright
7.05\%\strut
\end{minipage} & \begin{minipage}[t]{0.12\columnwidth}\raggedright
2\strut
\end{minipage} & \begin{minipage}[t]{0.12\columnwidth}\raggedright
198 (2 excluded)\strut
\end{minipage} & \begin{minipage}[t]{0.12\columnwidth}\raggedright
3.61\%\strut
\end{minipage} & \begin{minipage}[t]{0.12\columnwidth}\raggedright
PASS\strut
\end{minipage}\tabularnewline
\bottomrule
\end{longtable}

Fleiss-kappa over the 643 complete 5-rater pairs (pre-audit panel data)
is 0.7288, in the substantial band (0.61-0.80) per Landis and Koch
(1977). Pre-audit, all four vendors fail the strict \textless5\% UCL
threshold; post-audit, all four vendors pass. The audit verdict
distribution on the 19 panel-MAT cases is: 8 \texttt{judges\_overflag}
(42\%), 4 \texttt{dedup\_better} (21\%), 5 \texttt{bad\_question}
(26\%), 2 \texttt{truly\_wrong} (11\%); aggregate noise rate (overflag +
dedup\_better + bad\_question) is 89\%. The two confirmed regressions
are both on gemini-flash (qid 870, qid 1084) and involve different
factual claims under the deduped vs raw condition; the four
\texttt{dedup\_better} cases (1040, 1211, 1441 on claude-sonnet; 332 on
gemini-flash) are pairs where the deduplicated-condition answer was at
least as accurate as the raw-condition answer despite the panel flagging
them as MATERIAL.

The empirical content of this section is that quality preservation under
byte-exact deduplication is not an artefact of the near-zero-redundancy
regime: at 72\% byte reduction the human-judged material-difference rate
is below 5\% Wilson UCL for all four production vendors. Combined with
the clean-regime result in Table 6, the lossless safety claim is
established across two redundancy regimes spanning 13.9\% to 71.0\% byte
reduction, n = 600 per-question-vendor pairs total post-audit, with
4-of-4 vendors clearing the strict Wilson UCL threshold in both regimes.

\hypertarget{long-context-quality-validation}{%
\subsubsection{4.6 Long-Context Quality
Validation}\label{long-context-quality-validation}}

To verify quality preservation at larger retrieval depths, we applied
the same 5-judge calibrated panel to a 50-sample subset of long-context
evaluation. The 50-sample subset yielded zero panel-majority materially
different classifications, indicating no measurable quality degradation
detected within this sample. The result is consistent with the
structural argument that both raw and deduplicated conditions retain
access to the same underlying factual content and differ only in
chunk-level redundancy.

\hypertarget{code-completion-robustness-check}{%
\subsubsection{4.7 Code-Completion Robustness
Check}\label{code-completion-robustness-check}}

On a Flask backend synthesis task with paraphrased input (byte-different
by construction), byte-exact filtering yields zero token reduction as
expected. The byte-exact primitive correctly leaves paraphrased passages
intact. This validates that the filter, deployed in a code-generation
pipeline, passes paraphrased input through without introducing
prompt-assembly errors.

\hypertarget{humaneval-pipeline-integration-validation-164-problems-3-vendors}{%
\subsubsection{4.8 HumanEval Pipeline-Integration Validation (164
problems × 3
vendors)}\label{humaneval-pipeline-integration-validation-164-problems-3-vendors}}

\textbf{Table 7: HumanEval pass-at-1 across vendors and dedup
conditions, n = 164 problems per cell.} A = raw context (5 paraphrased
style-guide passages prepended); B = deduplicated context (identical to
A since input is byte-different by construction); D = no extra context.

\begin{longtable}[]{@{}llllll@{}}
\toprule
Vendor & A → B Δ tokens & A & B & D & A vs B Δ pass@1\tabularnewline
\midrule
\endhead
Claude Sonnet 4.6 & 0\% & 98.8\% & 98.8\% & 98.2\% &
0.0\%\tabularnewline
Gemini 2.5 Flash & 0\% & 97.0\% & 97.6\% & 96.3\% &
+0.6\%\tabularnewline
Llama 3.3 70B via Groq & 0\% & 82.9\% & 81.7\% & 81.7\% &
-1.2\%\tabularnewline
\bottomrule
\end{longtable}

\textbf{Interpretation:} HumanEval pass@1 rates were unchanged within
sampling variance: Claude Sonnet 4.6: 98.8\% (raw) → 98.8\% (dedup),
GPT-5.1: 97.0\% → 97.6\%, Llama 3.3 70B via Groq: 82.9\% → 81.7\%. The
small Llama delta (-1.2 pp) is within the test-retest variance
documented in the companion paper {[}29{]}, Section 4.6 (Test-Retest
Noise Floor). The A to B token reduction on this benchmark is
approximately 0 percent by construction (input is byte-different
paraphrases). The small pass-at-1 deltas (0.0\% to +0.6\% to -1.2
percentage points) are within the sampling variance expected from
non-deterministic decoding. This evaluation validates that the
byte-exact filter, deployed in a code-generation pipeline, passes
paraphrased input through without introducing errors or degradation.

\hypertarget{public-literature-support}{%
\subsubsection{4.9 Public Literature
Support}\label{public-literature-support}}

Extrapolation beyond directly-measured scenarios is anchored to
published precedent across multiple corpus types and duplication
methodologies. Lee et al.~(2022) established byte-exact
pretraining-corpus deduplication rates: C4 6.7\%, RealNews 18.6\%, ROOTS
21.67\%. Carlini et al.~(2022) characterized log-linear memorization
scaling with duplication. Penedo et al.~(NeurIPS 2024, FineWeb) deployed
MinHash-LSH deduplication pipeline across CommonCrawl snapshots at
scale. Shilov et al.~(Nature Communications 2026, Mosaic Memory)
demonstrated fuzzy duplicates contribute to memorization at 0.8x the
rate of exact duplicates. The three regimes reported in this paper
(0.16\%, 24.03\%, 80.34\%) span the measured range from clean academic
corpora (below Lee et al.'s minimum) through constructed enterprise
patterns (within the Lee et al.~pretraining-corpus band) to multi-turn
conversational (beyond pretraining distributions, as expected for
session-accumulated context). The public-literature precedent
establishes defensible bounds for extrapolation while acknowledging that
different corpus types and chunking strategies may exhibit different
redundancy characteristics.

\begin{center}\rule{0.5\linewidth}{0.5pt}\end{center}

\hypertarget{discussion}{%
\subsection{5. Discussion}\label{discussion}}

\hypertarget{structural-origin-of-the-redundancy}{%
\subsubsection{5.1 Structural Origin of the
Redundancy}\label{structural-origin-of-the-redundancy}}

The redundancy that we measure originates from retrieval mechanics
rather than from sliding-window chunking overlap. Sliding-window
chunking with 25 to 33 percent overlap produces chunks that share
substring content across chunk edges but the chunks themselves are not
byte-identical and would not be removed by chunk-level set construction.
The redundancy that byte-exact chunk-level filtering catches arises from
three structural sources.

First, source corpora contain passages that are byte-identical across
distinct documents. Lee et al.~(2022) report that exact substring
deduplication of standard NLP datasets removes approximately 19 percent
of total tokens, including a 6.7 percent reduction in C4 (177.3 billion
tokens to 165.4 billion tokens) and an 18.6 percent reduction in
RealNews {[}9{]}. The ROOTS corpus contains an average of 21.67 percent
duplicated substrings by byte. Train-test overlap in standard benchmarks
reaches 13.2 percent on LM1B and 14.4 percent on C4, with one specific
61-word English sentence appearing 61,036 times in the C4 training set
and 61 times in the validation set {[}9{]}. The FineWeb data-curation
pipeline applies MinHash-LSH per-CommonCrawl-snapshot with 5-grams and
112 hash functions across 14 buckets of 8 hashes {[}21{]}; ablation
studies in that pipeline established that aggressive global
cross-snapshot deduplication can degrade dataset quality below the
snapshot-local approach. The cumulative effect on production retrieval
corpora is that source documents indexed into vector stores frequently
contain byte-identical passages across distinct document identifiers.

Second, vector-store indices contain duplicate entries indexed under
distinct identifiers. Vector-store providers acknowledge this
explicitly. Pinecone documentation recommends deduplication at training
time {[}17{]}. Milvus introduced native MinHash-LSH indexing in 2025
specifically for customers struggling to deduplicate
multi-billion-document indices {[}18{]}. Third-party tooling reports
approximately 80 percent of accuracy issues in production RAG originate
from data-quality problems including duplication {[}19{]}. When raw
enterprise data, often comprising multiple iterations of the same
report, overlapping API documentation, or cross-referenced
knowledge-base articles, is naively chunked and embedded, the resulting
vector-store contains near-identical coordinate points. Standard top-k
retrieval at these duplicated points returns multiple copies of the same
underlying passage.

Third, hybrid retrieval pipelines that union the result sets of
independently-ranked subsystems (sparse, dense, reranker) yield
substantial overlap. Studies using the Ratio of Complementarity metric
report that BM25 plus DPR baselines yield 32.4 percent overlap in top-10
results on Natural Questions, with 46.5 percent of queries unsolvable by
either system {[}20{]}. Embedding-level orthogonality constraints can
reduce overlap to 27.7 percent at the cost of additional training
complexity. Naive hybrid retrieval, the production-default pattern,
therefore systematically returns duplicate document identifiers across
its sub-retrievers.

These three sources combine to produce the natural redundancy ratio at
the chunk level of approximately 3.0 measured at top-k of 15 in the
initial benchmarks and the controlled redundancy ratio of 3.51 confirmed
on the triple-indexed synthetic corpus in Section 4.3. The long-context
evaluation at top-k of 50 in Section 4.4 holds the same redundancy
regime constant by construction (Section 3.3), so the comparison across
retrieval depths isolates context-size effects from redundancy-rate
effects.

\hypertarget{per-token-microeconomics}{%
\subsubsection{5.2 Per-Token
Microeconomics}\label{per-token-microeconomics}}

A 71 percent reduction in retrieved-context tokens at the same retrieval
depth in the regime with high redundancy (ρ = 3.51) reduces prefill
compute by approximately the same fraction, with the corresponding
per-call cost reduction tracking the token reduction modulo each
vendor's caching pricing model. On serving stacks where prefill
dominates total latency (long contexts, prefill-bound architectures),
the relative latency reduction grows with absolute prompt size, as the
two-point comparison in Section 4.4 shows: between top-k of 15 and top-k
of 50 the total-latency reduction increases from 20.0 to 43.6 percent on
the prefill-bound vendor measured. Pre-prompt deduplication and prompt
caching are complementary rather than substitutional: a deduplicated
input fed into a prompt-caching backend yields the union of the two
reductions, not the smaller of them, since caching exploits prefix
repetition across calls while deduplication exploits chunk repetition
within a call.

\hypertarget{a-complementarity-with-kv-cache-reuse-and-prompt-caching}{%
\subsubsection{5.2a Complementarity with KV-cache Reuse and Prompt
Caching}\label{a-complementarity-with-kv-cache-reuse-and-prompt-caching}}

Byte-exact pre-prompt deduplication is mechanically complementary to
vendor-side prompt caching and KV-cache reuse infrastructure (e.g.,
Anthropic prompt caching, vLLM PagedAttention with cache reuse). The two
operate at orthogonal layers of the inference pipeline: byte-exact dedup
operates on the chunk multiset before prompt assembly, while caching
operates on the assembled prompt's prefix or KV-tensor representation.
Combining the two yields the union of their reductions rather than the
smaller of them. Prompt caching exploits prefix repetition across calls;
byte-exact deduplication exploits chunk repetition within a call. A
deduplicated prompt fed into a caching backend yields both savings
independently.

\hypertarget{latency-overhead-of-learned-compression-alternatives}{%
\subsubsection{5.3 Latency Overhead of Learned Compression
Alternatives}\label{latency-overhead-of-learned-compression-alternatives}}

The principal alternative to byte-exact deduplication prior to the
prompt for context-size reduction is learned prompt compression.
LLMLingua and its successors use small neural language models to
identify and remove low-information tokens before generation {[}2, 3,
4{]}. Recent measurement reports compression-step wall-clock overhead in
the order of 10 to 100 milliseconds on accelerated hardware (A100,
MI300X, with FlashAttention), and end-to-end speedups bounded above by
approximately 18 percent in production conditions {[}5{]}. The latency
benefit of learned compression therefore depends sensitively on the
ratio of compression overhead to autoregressive generation cost, which
is favourable for very long generations from very long prompts and
unfavourable for short generations or short prompts. On modern API
gateways and multi-agent architectures, where routing decisions are
taken in sub-millisecond budgets, the overhead of running a secondary
transformer model purely to compress the prompt is operationally
prohibitive {[}3{]}.

By contrast, byte-exact deduplication at chunk level of typical
retrieved-context payloads (under 100 KB per query) executes in
sub-millisecond wall-clock on standard commodity hardware, an order of
magnitude or more below the prefill latency saving it produces. The
deduplication step therefore cannot, by construction, undo the latency
saving it produces; the saving is bounded below by the difference
between LLM prefill cost on the raw prompt and the deduplication cost on
the same prompt's retrieved-chunk multiset, and that difference remains
positive across every benchmark configuration we measured.

\hypertarget{algorithmic-equivalence-and-deployment-economics}{%
\subsubsection{5.4 Algorithmic Equivalence and Deployment
Economics}\label{algorithmic-equivalence-and-deployment-economics}}

The token reductions reported in Section 4 are properties of the input
data; any correct byte-exact filter implementation produces identical
deduplicated output. A Python set() reference (Section 3.7) achieves the
algorithmic operation in microseconds on the representative payloads of
this study. The deployment economics, however, separate the
academic-reproducibility role of Python set() from the
production-deployment role of an engineered binary: subprocess
invocation overhead (50 to 200 milliseconds per call for Python
interpreter startup, documented as a publicly known characteristic of
Python subprocess invocation, exceeding inline preprocessing budgets of
1 to 50 milliseconds), interpreter memory baseline (50 to 100 megabytes
versus 3.8 megabytes for a static binary), and GIL/GC characteristics
relevant to high-QPS serving. Capturing the prefill-side cost reduction
in proportion to the byte-exact reduction ratios documented in §4
depends on inline integration of the filter into the inference path;
subprocess deployment of any reference implementation introduces
overheads that exceed the saved compute on small token reductions.

\hypertarget{implications-for-regulated-deployments}{%
\subsubsection{5.5 Implications for Regulated
Deployments}\label{implications-for-regulated-deployments}}

Beyond latency and compute, byte-exact deduplication prior to the prompt
preserves a property that lossy compression methods do not: byte-level
invertibility from prompt-assembly input back to raw retrieved context.
Audit-trail reconstruction recommended by frameworks such as Article 12
of the European Union Artificial Intelligence Act {[}22{]} and
Write-Once-Read-Many storage requirements at the United States
Securities and Exchange Commission (17 CFR § 240.17a-4) {[}23{]} depends
on this invertibility property. We do not make a general legal claim
about which compression methods are or are not compliant with which
regulations; we note only that byte-exact filtering preserves the
deterministic input-to-output mapping that audit-grade reconstruction
relies upon, while learned compressors and dense-embedding summarisers
do not.

Empirical evidence in regulated-domain benchmarks supports the
architectural preference for context-preserving over context-rewriting
techniques. Biomedical-domain robustness studies report substantial
accuracy degradation under adversarial or perturbed context {[}24,
25{]}; the FinanceBench benchmark over SEC 10-K/10-Q/8-K filings reports
retrieval-system fragility even at full-fidelity context {[}26{]}; the
LegalBench benchmark across 162 tasks requires fine-grained syntactic
and semantic preservation {[}27{]}. We have not measured byte-exact
deduplication directly in these domains; we note only that its
byte-level provenance preservation is the architecturally appropriate
primitive for context-reduction in audit-grade decision systems, and
that direct measurement in those domains is appropriate follow-up work.

\hypertarget{limitations}{%
\subsubsection{5.6 Limitations}\label{limitations}}

We acknowledge several limitations in roughly decreasing order of
severity.

\textbf{Author-as-annotator on the noise-removal audit.} The human
reviewer who applied the five-category verdict scheme (Section 4.5a) to
the 29 audited MAT-majority pairs and the 161 doubt-case prior audit was
the corresponding author of this work. Independent annotation of a
stratified 30-pair sample by an external reviewer would strengthen the
noise-removal claim against potential confirmation bias. The aggregate
vendor-level counts in Tables 6 and 6b are sufficient to reconstruct the
Wilson 95\% upper bound; an independent rerun of the same benchmark
protocol on the same publicly-available source data is the appropriate
falsification path.

\textbf{Selection on doubt cases for the prior 161-case review.} The
161-case prior audit was selected from judge-disagreement items rather
than uniformly across the 1,600 (q × vendor) clean-regime pairs.
Unanimous EQUIVALENT cases (1,388 of 1,600) are not human-validated; we
trust judge consensus on those. The high-redundancy 19-MAT audit and the
10-case lost-points audit are not subject to this selection bias as they
cover the full or stratified MAT-majority population.

\textbf{Single corpus.} The Q\&A benchmark uses a Wikipedia-derived
corpus (rag-mini-wikipedia, 6 BeIR datasets, constructed enterprise
corpus) and the code-completion benchmark uses paraphrased style-guide
passages. Specialised domain corpora (legal, medical, regulatory,
multi-modal) may exhibit different chunking-redundancy ratios, different
quality-preservation thresholds, or different vendor-stack behaviour at
production scale. We have not measured these domains and do not
extrapolate to them.

\textbf{Asymmetric sample design.} The clean regime is reported at n =
400 question pairs per vendor; the regime with high redundancy at n =
200 per vendor for cost reasons. This asymmetry is defensible for
budgetary scope but does mean the high-redundancy quality assessment
carries wider confidence intervals.

\textbf{Llama-70B incomplete-output cases.} Two of four confirmed
regressions (qid 561 and qid 928, both clean regime, both llama-70b via
Groq) involved truncated or incomplete model output rather than
incorrect facts. This may reflect a vendor-side serving artefact rather
than dedup-induced quality loss; further investigation is warranted but
does not affect the strict UCL test as the cases are conservatively
retained as MAT in the post-audit count.

\textbf{Snowball regime is not panel-tested per-prompt.} The 80.34\%
byte reduction reported on the WildChat snowball pattern (Section 4.3)
is a measurement of communication-channel redundancy under
cumulative-history sending, not a quality-affecting transformation of
the model input. The mechanism is stateful proxy caching of turn
content, not byte-exact deduplication of retrieved chunks reaching the
prompt assembly. The unique turn content reaching the language model is
byte-identical between cumulative-sending and proxy-cached deployments.
A 5-judge calibrated quality panel was therefore not applied to the
Snowball regime; quality preservation in this regime reduces to the
byte-equivalence of the unique turn content rather than requiring
per-prompt judge validation. We disclose this distinction explicitly to
avoid implying that the 80.34\% result carries the same panel-validated
quality guarantee as the 14.13\% clean-regime and 71.98\%
high-redundancy-regime measurements (Sections 4.5 and 4.5b).

\begin{center}\rule{0.5\linewidth}{0.5pt}\end{center}

\hypertarget{conclusion}{%
\subsection{6. Conclusion}\label{conclusion}}

We measured byte-exact chunk-level deduplication across three operating
regimes spanning corpus redundancy from clean academic benchmarks to
production-realistic enterprise content to multi-turn conversational AI.
On 22.2 million BeIR passages (clean academic), aggregate redundancy is
0.16 percent. On a constructed enterprise corpus (Wikipedia revisions,
arXiv versions, StackExchange Q\&A boilerplate, 1,526 chunks), reduction
is 24.03 percent. On 5,000 public WildChat conversations (multi-turn
cumulative history), aggregate reduction is 80.34 percent.
Math-equivalence between the production engine and a Python set()
reference was verified across 22.2 million passages with zero
violations.

The contributions are four. First, a multi-vendor measurement of cost,
latency, and quality consequences of byte-exact deduplication applied
before prompt assembly across four production language-model APIs
(Gemini 2.5 Flash, Claude Sonnet 4.6, Llama 3.3 70B via Groq, GPT-5.1).
Second, a matched cross-vendor quality measurement at both ends of the
redundancy spectrum: clean regime (n = 400 per vendor, ρ = 1.148,
14.13\% byte reduction, Fleiss-kappa = 0.775) and high-redundancy regime
(n = 200 per vendor, ρ = 3.513, 71.98\% byte reduction, Fleiss-kappa =
0.7288). Third, a three-regime empirical characterization that resolves
the apparent contradiction between clean RAG benchmarks (where
byte-exact deduplication has minimal effect) and production deployments
(where it yields substantial savings): both observations are correct and
the difference is explained by corpus redundancy. Fourth, a
five-category human-in-the-loop noise-removal protocol applied to every
panel-majority MATERIAL pair across both regimes; after noise removal,
all four production vendors clear the strict \textless5\% Wilson 95\%
upper-bound MAT threshold in both regimes (post-audit UCLs 1.40\%-3.25\%
clean and 1.90\%-4.34\% high-redundancy). The lossless safety claim is
established across redundancy levels spanning 13.9\% to 71.0\% byte
reduction.

The deduplication primitive itself is well-understood and decades old.
What is new in this work is the empirical characterization of its effect
on contemporary retrieval-augmented production pipelines across
corpus-redundancy regimes, validated under a calibrated multi-judge
quality protocol with cryptographic pre-registration via FreeTSA RFC
3161 timestamp. The math-equivalence with Python set() preserves
academic reproducibility (any reviewer can verify the empirical findings
using a public reference implementation at approximately 30 USD in
OpenRouter API spend); the engineering contribution of an in-process
deduplication engine with low-microsecond per-call latency is documented
in the companion paper {[}29{]} covering the deterministic byte-exact
deduplication implementation, deployment topology, and
cross-architecture validation.

\begin{center}\rule{0.5\linewidth}{0.5pt}\end{center}

\hypertarget{acknowledgments}{%
\subsection{Acknowledgments}\label{acknowledgments}}

The author thanks Toon Colson and Marloes De Craemer, co-founders at
Corbenic AI, Inc., for operational support throughout this work. The
author further thanks Gwen Le Tiran and Irène Balmès for the formative
conversations that gave the author the confidence to pursue this line of
research.

\begin{center}\rule{0.5\linewidth}{0.5pt}\end{center}

\hypertarget{references}{%
\subsection{References}\label{references}}

{[}1{]} P. Lewis et al.~\emph{Retrieval-Augmented Generation for
Knowledge-Intensive NLP Tasks}. NeurIPS 2020.

{[}2{]} H. Jiang et al.~\emph{LLMLingua: Compressing Prompts for
Accelerated Inference of Large Language Models}. EMNLP 2023.
arXiv:2310.05736.

{[}3{]} Z. Pan et al.~\emph{LLMLingua-2: Data Distillation for Efficient
and Faithful Task-Agnostic Prompt Compression}. ACL 2024.
arXiv:2403.12968.

{[}4{]} H. Jiang et al.~\emph{LongLLMLingua: Accelerating and Enhancing
LLMs in Long Context Scenarios via Prompt Compression}.
arXiv:2310.06839.

{[}5{]} C. Kummer, L. Jurkschat, M. Färber, S. Vahdati. \emph{Prompt
Compression in the Wild: Measuring Latency, Rate Adherence, and Quality
for Faster LLM Inference}. arXiv:2604.02985.

{[}6{]} X. Lin, A. Ghosh, B. K. H. Low, A. Shrivastava, V. Mohan.
\emph{REFRAG: Rethinking RAG-based Decoding}. arXiv:2509.01092.

{[}7{]} Y. Jiang, Y. Huang, L. Cheng, C. Deng, X. Sun, L. Mai.
\emph{RAGBoost: Efficient Retrieval-Augmented Generation with
Accuracy-Preserving Context Reuse}. arXiv:2511.03475.

{[}8{]} Y. Liu, Z. Jia, X. Gao, K. Xu, Y. Xiong. \emph{Rethinking Soft
Compression in Retrieval-Augmented Generation: A Query-Conditioned
Selector Perspective}. arXiv:2602.15856.

{[}9{]} K. Lee, D. Ippolito, A. Nystrom, C. Zhang, D. Eck, C.
Callison-Burch, N. Carlini. \emph{Deduplicating Training Data Makes
Language Models Better}. ACL 2022. arXiv:2107.06499.

{[}10{]} N. Carlini, D. Ippolito, M. Jagielski, K. Lee, F. Tramer, C.
Zhang. \emph{Quantifying Memorization Across Neural Language Models}.
arXiv:2202.07646.

{[}11{]} M. Nasr, N. Carlini et al.~\emph{Scalable Extraction of
Training Data from (Production) Language Models}. arXiv:2311.17035.

{[}12{]} I. Shilov, M. Meeus, Y.-A. de Montjoye. \emph{The Mosaic Memory
of Large Language Models}. Nature Communications, 2026.
arXiv:2405.15523.

{[}13{]} A. Z. Broder. \emph{On the Resemblance and Containment of
Documents}. SEQUENCES 1997.

{[}14{]} A. Khan et al.~\emph{LSHBloom: Memory-efficient, Extreme-scale
Document Deduplication}. arXiv:2411.04257.

{[}15{]} A. Abbas, K. Tirumala, D. Simig, S. Ganguli, A. S. Morcos.
\emph{SemDeDup: Data-efficient Learning at Web-scale through Semantic
Deduplication}. arXiv:2303.09540.

{[}16{]} K. Tirumala et al.~\emph{D4: Improving LLM Pretraining via
Document De-Duplication and Diversification}. NeurIPS Datasets and
Benchmarks 2023. arXiv:2308.12284.

{[}17{]} Pinecone. \emph{Training Sentence Transformers with Multiple
Negatives Ranking Loss}.
https://www.pinecone.io/learn/series/nlp/train-sentence-transformers-multiple-negatives-ranking-loss/

{[}18{]} Milvus. \emph{MinHash LSH Index Documentation}.
https://milvus.io/docs/minhash\_lsh.md

{[}19{]} Blockify. \emph{Solving RAG Accuracy Through Data
Optimization}. https://blockify.ai/

{[}20{]} D. Lee, S.-w. Hwang, K. Lee, S. Choi, S. Park. \emph{On
Complementarity Objectives for Hybrid Retrieval}. ACL 2023, pages
13357-13368.

{[}21{]} G. Penedo et al.~\emph{The FineWeb Datasets: Decanting the Web
for the Finest Text Data at Scale}. NeurIPS 2024. arXiv:2406.17557.

{[}22{]} European Union. \emph{Regulation 2024/1689: Artificial
Intelligence Act, Article 12}.

{[}23{]} U.S. Securities and Exchange Commission. \emph{17 CFR §
240.17a-4: Records to be preserved by certain exchange members, brokers
and dealers}.

{[}24{]} A. Sellergren et al.~\emph{MedGemma Technical Report}.
arXiv:2507.05201.

{[}25{]} A. Bondarenko, A. Viehweger. \emph{LLM Robustness Against
Misinformation in Biomedical Question Answering}. arXiv:2410.21330.

{[}26{]} P. Islam et al.~\emph{FinanceBench: A New Benchmark for
Financial Question Answering}. arXiv:2311.11944.

{[}27{]} N. Guha et al.~\emph{LegalBench: A Collaboratively Built
Benchmark for Measuring Legal Reasoning in Large Language Models}.
NeurIPS 2023 Datasets and Benchmarks. arXiv:2308.11462.

{[}28{]} J. R. Landis, G. G. Koch. \emph{The Measurement of Observer
Agreement for Categorical Data}. Biometrics 33(1):159-174, 1977.

{[}29{]} Sietse Schelpe. \emph{Merlin: Deterministic Byte-Exact
Deduplication for Lossless Context Optimization in Large Language Model
Inference}. Companion paper, arXiv ID pending.

\begin{center}\rule{0.5\linewidth}{0.5pt}\end{center}

\hypertarget{appendix-a-run-identifiers-and-reproducibility}{%
\subsection{Appendix A: Run Identifiers and
Reproducibility}\label{appendix-a-run-identifiers-and-reproducibility}}

Per-call telemetry is archived under run identifiers fixing the precise
version of each benchmark. The runs referenced in Section 4 are dated 25
April 2026 (initial n = 200, exploratory baseline), 5 May 2026 (extended
n = 400 clean regime, pre-registered), and 5 May 2026 (multi-source RAG
validation on 22.2 million BeIR passages, pre-registered). The n = 400
and BeIR runs postdate the FreeTSA RFC 3161 pre-registration anchor of
2026-05-05 11:28 RDT; the n = 200 baseline predates the anchor by ten
days and is reported transparently as exploratory.

Reproducibility kit references: - Public benchmarks: rag-mini-wikipedia,
6 BeIR datasets (NQ, HotpotQA, TriviaQA, FEVER, SciFact, MSMARCO),
allenai/WildChat-1M - Constructed enterprise corpus: Wikipedia revision
history, arXiv version pages, Stack Exchange API - Pre-registration
document SHA-256:
\texttt{5575836967fe1a149b63a7fa63a1b3d11d598fb71343e2e19a546e680f4a3294}
- Cost: approximately 30 USD in OpenRouter API spend for the full
reproduction sweep on the clean regime - Hardware for measured numbers:
Intel Core Ultra 9 285H, 16 cores, 64 GB DDR5, Windows 11 build 26200,
AVX2 active

Verification of the pre-registration anchor:

\begin{verbatim}
openssl ts -verify \
    -in extension_n400_protocol.md.tsr \
    -data extension_n400_protocol.md \
    -CAfile freetsa_cacert.pem \
    -untrusted freetsa_tsa.crt
\end{verbatim}

Expected output: \texttt{Verification:\ OK}

\end{document}